\documentclass{IEEEtran}
\usepackage{amsmath,amssymb,amsfonts}
\usepackage{mathtools}
\usepackage{dsfont}
\usepackage{hyperref}
\usepackage{algorithmic}
\usepackage{graphicx}
\usepackage{textcomp}
\usepackage{subfigure}
\usepackage{tabu} 
\usepackage{multirow}
\usepackage{hyperref}
\hypersetup{
colorlinks=true,
urlcolor=black
}
\usepackage{cite}
\usepackage{color}
\usepackage[dvipsnames, svgnames, x11names]{xcolor}
\usepackage{algorithm}
\usepackage{algorithmic}
\usepackage[]{footmisc}
\usepackage{booktabs}       
\usepackage{bm}
\usepackage{threeparttable}
\usepackage{colortbl}
\usepackage{diagbox}
\usepackage{tikz}

\begin{document}
\title{Emotion Separation and Recognition from a Facial Expression by Generating the Poker Face with Vision Transformers}
\title{Emotion Separation and Recognition from a Facial Expression by Generating the Poker Face with Vision Transformers}
\author{Jia Li, Jiantao Nie, Dan Guo, Richang Hong\textsuperscript{*}, Meng Wang\textsuperscript{*}, \IEEEmembership{Fellow, IEEE}
\thanks{The authors are all with the School of Computer Science and Information Engineering, Hefei University of Technology, Hefei  230009, China (e-mail: jiali@hfut.edu.cn; jiantaonie@mail.hfut.edu.cn; guodan@hfut.edu.cn; hongrc.hfut@gmail.com; eric.mengwang@gmail.com). (\emph{*Corresponding authors: Richang Hong, Meng Wang}.)}
\thanks{This work was supported by the National Natural Science Foundation of China under Grants 62202139 and 61932009, and in part by The University Synergy Innovation Program of Anhui Province under Grant GXXT-2022-038. Special thanks to Ziyang Zhang for his previous suggestions and help on running experiments.
}
\thanks{©2024 IEEE. Personal use of this material is permitted. Permission from IEEE must be
obtained for all other uses, in any current or future media, including
reprinting/republishing this material for advertising or promotional purposes, creating new
collective works, for resale or redistribution to servers or lists, or reuse of any copyrighted
component of this work in other works. \\
Digital Object Identifier: 10.1109/TCSS.2024.3478839}
}


\maketitle

\begin{abstract}

Representation learning and feature disentanglement have garnered significant research interest in the field of facial expression recognition (FER).  The inherent ambiguity of emotion labels poses challenges for conventional supervised representation learning methods. Moreover, directly learning the mapping from a facial expression image to an emotion label lacks explicit supervision signals for capturing fine-grained facial features. 
In this paper, we propose a novel FER model, named Poker Face Vision Transformer or PF-ViT, to  address these challenges. PF-ViT aims to separate and recognize the disturbance-agnostic emotion from a static facial image via generating its corresponding poker face, without the need for paired images. Inspired by the Facial Action Coding System, we regard an expressive face as the  combined result of a set of facial muscle movements on one's poker face (i.e., an emotionless face). PF-ViT utilizes vanilla Vision Transformers, and its components are firstly pre-trained as Masked Autoencoders on a large facial expression dataset without emotion labels, yielding excellent representations. Subsequently, we train PF-ViT using a GAN framework. During training, the auxiliary task of poke face generation promotes the disentanglement between emotional and emotion-irrelevant components, guiding the FER model to holistically capture discriminative facial details.
Quantitative and qualitative results demonstrate the effectiveness of our method, surpassing the state-of-the-art methods on four popular FER datasets. 

\end{abstract}

\begin{IEEEkeywords}
Facial expression recognition, representation learning, vison transformer, generative adversarial network.
\end{IEEEkeywords}

\section{Introduction} \label{sec:introduction}
Humans typically exhibit non-verbal emotions through their facial expressions, whether consciously or unconsciously.
Given a single facial image, the task of facial expression recognition (FER)  aims at recognizing the basic expression on the human face \cite{zhang2022learn,yang2018facial,zhou2024seeing,zhao2021robust,ma2023transformer,wang2024graph}, such as happiness, sadness, and contempt, which can facilitate many applications in computer vision, mental health assessment, etc \cite{li2024automatic,li2022deep,liu2024multimodal}. Predicting the discrete emotion category from a static facial image in the wild is quite challenging due to several factors, such as the inherent ambiguity of expression, occlusion and head pose variation\cite{zeng2022face2exp,zhang2022learn}. Other intertwined irrelevant factors, such as  background illumination, identity bias, and imbalanced, noisy emotion labels \cite{xue2021transfer,zhang2021joint,jiang2022disentangling,wang2019identity} in real-world FER datasets make the problem of in-the-wild FER more difficult to address.

\begin{figure}
	\centerline{\includegraphics[width=0.85\linewidth]{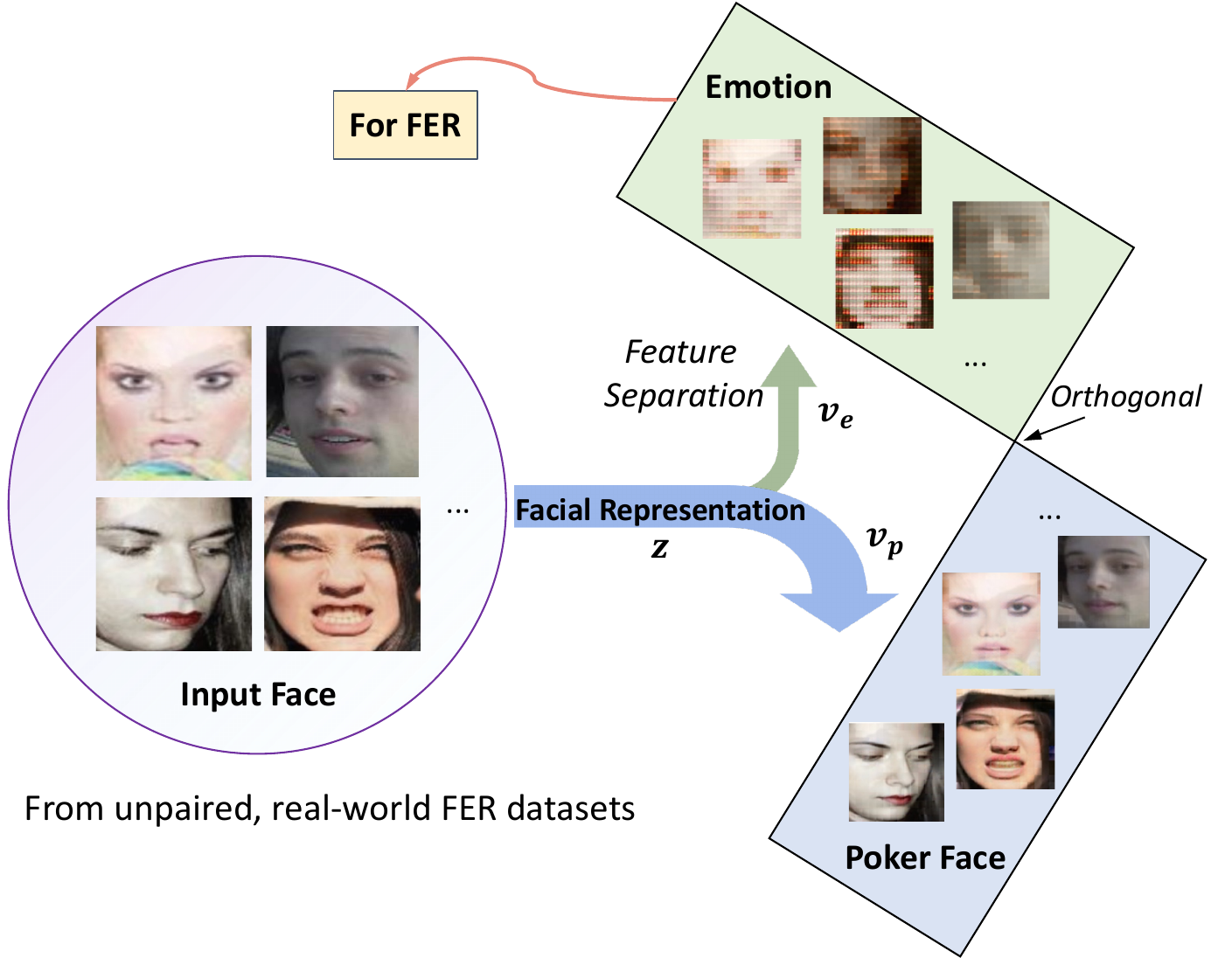}} 
	   \caption{Illustration of emotion separation from expressive faces in a latent space. We assume that an expressive face is the combined outcome of a set of facial muscle movements on one’s poker face. Our method focuses on separating the emotional component and preserving all  emotion-irrelevant details (i.e., disturbance) to synthesize the emotionless counterpart.}
	\label{fig:separation}
	\end{figure}

A \emph{poker face} refers to a deliberately-induced blank expression on the face of a person that does not show any thinking or feeling of that person.
The well-known Facial Action Coding System (FACS) \cite{ekman1978facial,lien1998automated,zhang2021learning}, studied and developed for many decades,  describes a facial expression as the combination of a set of distinct local action units (AUs).  In other words, a facial expression can be roughly decomposed to an expressive component and a neutral component \cite{wang2003facial,calder2005understanding,yang2018facial}.  Building upon this understanding, \emph{we consider an expressive face as the comprehensive result of a set of facial muscle movements on someone's poker face. Here, \textbf{we make an assumption that the face displaying neutral emotion is nearly emotionless and is equal to a poker face}}. Neutral expression and happy expression are the two most frequently occurring expressions in popular real-world FER datasets, such as AffectNet \cite{mollahosseini2017affectnet}, RAF-DB \cite{li2017reliable} and FERPlus \cite{barsoum2016training}.

Recently, researches have made substantial progress in the field of FER and transfer their research focus from lab-controlled settings to in-the-wild circumstances, benefiting much from the development of deep learning techniques and the availability of more representative FER datasets \cite{mollahosseini2017affectnet,li2017reliable}. However, it is important to note that deep neural networks typically require a large amount of training data, and the presence of noisy emotion labels can significantly hinder their generalization performance.
To address this challenge, many state-of-the-art (SOTA)  methods \cite{xue2021transfer,li2021mvt,she2021dive,hwang2022vision,wang2020suppressing} adopt a two-step approach. They first pre-train their deep models on a large scale and less noisy dataset, for instance, Ms-Celeb-1M \cite{guo2016ms} for face recognition (FR)\footnote{However, this large-scale face recognition dataset consisting of 10 million facial images has already been retracted and should not be used.}, in order to gain better facial representations and then fine-tune the deep models on specific FER datasets.
Various network architectures have been explored for FER, including Convolutional Neural Networks (CNNs), Graph Convolutional Networks (GCNs) and some hybrid Vision Transformers (hybrid ViTs) \cite{li2021mvt,ma2021facial,xue2021transfer,xue2022vision,li2023fg}. A hybrid ViT,  in particular, consists of a pre-trained CNN backbone and multiple Transformer \cite{vaswani2017attention} blocks. CNN-based FER models still dominate this field. SCN  \cite{wang2020suppressing}, RUL \cite{zhang2021relative}, and EAC \cite{zhang2022learn}  adopt the ResNet-18 model \cite{he2016deep} pre-trained on the MS-Celeb-1M dataset \cite{guo2016ms} as the backbone to extract facial features. 
Additionally, some advanced methods leverage off-the-shelf facial knowledge such as facial landmarks \cite{zhang2017facial} or hand-crafted image features \cite{ma2021facial} to aid their FER models for higher evaluation metrics.

Compared with CNNs, vanilla ViTs \cite{dosovitskiy2021an,steiner2021train} need much more training data to obtain excellent feature representations, for they use weaker inductive biases and capture more long-range dependency relationships \cite{she2021dive,zhang2021learning}. \emph{Taking into account that different facial expressions can share similar local appearences (i.e., facial muscle movements), such global attribute of Transformers may help alleviate the common overfitting issue in FER}. This issue arises when the model tends to memorize local facial features instead of learning from the global features. 
TransFER \cite{xue2021transfer} and VTFF \cite{ma2021facial} employ ResNet-based hybrid ViTs to perform the task and achieve competitive FER performance. FG-AGR \cite{li2023fg} introduces a fine-grained association graph representation framework for FER, combining CNN, GCN and Transformer simultaneously. 
Inspired by Generative Adversarial Networks (GANs), MVT \cite{li2021mvt} proposes a pure Transformer-based framework called Mask Vision Transformer, which aims to filter out the background and occlusion in face images.
It is worth noting that the current methods based on vanilla (pure) ViTs still lag behind those using CNNs or hybrid ViTs in terms of FER accuracy, model size and pre-training cost.  Consiquently, we raise the first question.

\textbf{Question \texttt{\#}1: \emph{Can vanilla ViTs, which capture long-range relationships using the naive global attention mechanism,  achieve outstanding FER performance without extra training data from upstream tasks?}} This problem has not been well explored thus far, and our answer is yes.

In addition to the network structure design and representation learning methods, identity-invariant FER has attracted more and more attention most recently \cite{wang2019identity,zhang2021learning,jiang2022disentangling,zhang2021joint}. 
Previous studies  \cite{wang2019identity,zhang2021learning,jiang2022disentangling} have attempted to disentangle facial emotion apart from person identity and even head pose, leveraging pre-trained FR models \cite{schroff2015facenet,kim2022adaface} or direct supervision of subject labels. CNN-based GANs are employed to generate new facial expressions \cite{zhang2021joint,xie2020facial}, thereby encouraging the FER model to learn a disentangled expressional representation and also enlarging the FER training set. However, other irrelevant factors except identity and pose still exist as before and are detrimental to FER. Some novel identity-aware FER methods \cite{huang2021identity,yang2018facial} essentially aim to  mitigate the adverse influence of personal identity as well. Unfortunately, these existing identity-invariant or identity-aware methods either need paired facial expression images for each subject or can not generate high-quality facial expressions for unseen subjects. In-the-wild FER datasets, such as AffectNet \cite{mollahosseini2017affectnet} and RAF-DB \cite{li2017reliable}, contain a large number of person identities and consist of  unpaired facial images only. Hence, we proceed to raise the second question.

\textbf{Question \texttt{\#}2: \emph{Can we use ViT-based GAN to simultaneously separate pure emotion features for higher FER accuracy and remove emotion from unconstrained facial expression images without the need for paired images?}} Our answer to this question is also positive.

However, training a plain ViT-based GAN (ViTGAN) for image generation is a challenging task  \cite{leevitgan,niu2022four}. And how to make the de-expression (or emotion disentanglement) task \cite{yang2018facial,xie2020facial,zhang2021learning} enhance FER remains to be further studied.
Masked autoencoder (MAE) \cite{he2022masked}, which reconstructs the missing parts of an input image using an asymmetric encoder-decoder architecture, is a very powerful self-supervised representation learing framework. MAEs have proven to be effective in various tasks like image inpainting and denoising. They can then be utilized in downstream tasks such as image generation, classification, and feature extraction.
Inspired by MAE, we transform and develop a vanilla ViT-based Masked Autoencoder into a GAN framework to accomplish a specific image-to-image translation task, which is to generate the emotionless counterpart, a poker face, given a facial image with an arbitrary expression. 
As illustrated in Fig. \ref{fig:separation}, our work explores how to separate the emotion-related features from the emotion-irrelevant features in a latent representation space through the auxiliary task of adversarial poker face generation. By introducing the auxiliary task, our aim is to fully disentangle the emotion details from the expressive face and compel the FER model to capture more fine-grained features of facial muscle movements, leading to a better understanding of facial expressions.

Motivated by the analysis above, we propose a plain ViT-based GAN model, primarily developed from MAEs \cite{he2022masked}. Our GAN model, referred to as \emph{\textbf{P}oker \textbf{F}ace \textbf{M}AE-\textbf{G}AN (\textbf{PF-MAG})}, is  depicted in Fig. \ref{fig:PFViT}. After training, we can take out a multi-task ViT model from PF-MAG capable of simultaneously perfroming facial expression classification and poker face generation. And we name this ViT model as \emph{\textbf{P}oker \textbf{F}ace \textbf{V}ision \textbf{T}ransformer (\textbf{PF-ViT})}.
PF-ViT mainly incorporates an image encoder, a token separator, an image generator, and a facial expression classification head. PF-ViT and a composite discriminator constitute our GAN model, in which the encoder, generator and discriminator are pure ViTs. Specifically, our method firstly trains the ViT encoder to learn fine-grained, robust facial expression representations using MAE pre-training \cite{he2022masked} on a large facial expression dataset without emotion labels.
After that, PF-ViT is trained with the discriminator in an adversarial manner to separate the discriminative emotion features from the learned representations and synthesize realistic, emotionless poke faces using its generator, deceiving the discriminator within the GAN framework. Meanwhile, FER supervision signals applied to the classification head of PF-ViT prevent the information collapse in the pure emotion representations separated by the token separator.
During testing, the generator of PF-ViT can be removed, and let PF-ViT solely perform the FER task.

Our main contributions are summarized as follows:
\begin{itemize}  
	\item We propose a novel in-the-wild FER model called PF-ViT, which utilizes plain ViTs to separate and recognize facial expressions and generate corresponding emotionless faces simultaneously, given a facial image. We train PF-ViT in a GAN framework without the need for paired images, where PF-ViT is encouraged to disentangle the pure emotion-related features, leading to an obvious improvement in FER performance.
	\item  We demonstrate the high potential of vanilla ViTs for FER by employing the MAE representation learing framework to pre-train ViT encoders using only unlabeled FER data. Furthermore, we show a tiny ViT with only 5.6M parameters has achieved decent performance, compared to previous advanced FER models\footnote{These MAE pre-training models will be available at: \url{https://github.com/MSA-LMC/MAE-SFER}.}.  
	\item Our method achieves SOTA performance on popular in-the-wild FER datasets, obtaining the highest accuracy of 92.07\%, 67.23\%, 64.10\% and 91.16\%  on RAF-DB, AffectNet-7, AffectNet-8 and FERPlus datasets, respectively.  Moreover, to the best of our knowledge, this is the first attempt to synthesize a realistic emotionless face from an arbitrary expressive face using plain Transformer backbones without paired data. 
\end{itemize}

The remainder of this paper is organized as follows. Sec. \ref{sec:related-work} reviews the related work briefly. Details of our method are elaborated in Sec. \ref{sec:our-method}. Sec. \ref{sec:our-experiment} provides the experimental results and analysis. Finally, we conclude this paper in Sec. \ref{sec:our-conclusion}.

\begin{figure}[!h]
	\centerline{\includegraphics[width=0.85\linewidth]{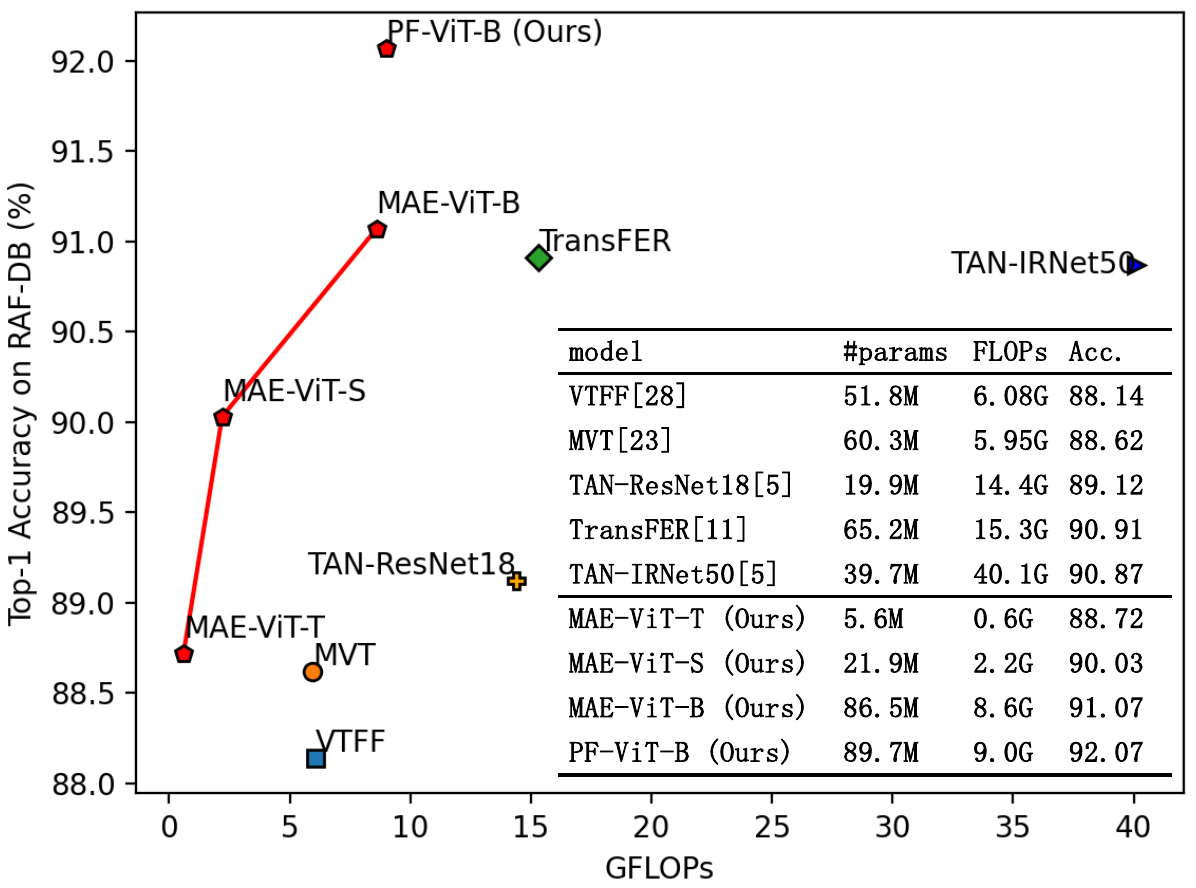}}
		\caption{Accuracy vs. GFLOPs during inference --  Comparison of our FER models with SOTA FER models on the RAF-DB testing set \cite{li2017reliable}  regarding accuracy, computational complexity (GFLOPs) during tesing, and model size (\#params). Our models use ViT-Base, ViT-Small and ViT-Tiny \cite{touvron2021training} as the image encoders, with an input size of $224 \times 224$. In this analysis, our proposed PF-ViT model utilizing ViT-Base as the image encoder is denoted as PF-ViT-B, and its image generator and cross-fusion module are not included during this testing.
		}
	\label{fig:FLOPs}
	\end{figure}

\section{Related Work} \label{sec:related-work}
Over the past few decades, extensive efforts have been devoted to addressing the problem of facial expression recognition (FER). Classical approaches concentrate on this problem under lab-controlled conditions and rely on handcrafted features \cite{ekman1978facial,lien1998automated}.
With the rapid advancements in deep neural networks and representation learning, substantial progress has been achieved and robust FER in real-world scenarios has become feasible \cite{xue2021transfer,she2021dive,li2021mvt,wang2020suppressing}, as discussed in Sec. \ref{sec:introduction}. In this section, we briefly review the most relevant studies on feature disentanglement and representation learning for FER.

\subsection{Feature Disentanglement for FER}  
To mitigate the disturbance caused by emotion-irrelevant factors, such as age, race, gender, illumination and occlusion, explicit and implicit feature disentanglement has been introduced into this field to extract disentangled facial expression representations.  DDL \cite{ruan2020deep} proposes a disturbance-disentangled learning  method  to disentangle several disturbing factors simultaneously.  FDRL \cite{ruan2021feature} presents a feature decomposition and reconstruction learning method that models expression similarities and expression-specific variations, enabling the extraction of fine-grained expression features.  To approach the goal of identity- and pose-robust FER, IPFR \cite{wang2019identity} employs adversarial feature learning to extract expression features while avoiding subject and pose variations. Subsequently, IPD-FER \cite{jiang2022disentangling} introduces a GAN-based FER model to disentangle identity, pose and expression. IPD-FER employs three independent CNNs to encode identity, pose and expression information, respectively. 
However, these methods requires the dataset with multi-task labels, which are unavailable for in-the-wild FER datasets.

Zhang et al. \cite{zhang2021joint} and Xie et al. \cite{xie2020facial}  employ CNN-based GANs to disentangle expressions and even generate new faces with different expressions (i.e., expression transfer) to enlarge the training set. However, the synthesized facial expressions lack realism and their FER performance falls short of expectations, indicating that the intrinsic expressional information is not effectively separated. 
By contrast, DeRL \cite{yang2018facial}, a two-phase method, firstly trains a conditional GAN (cGAN) model to generate the corresponding neutral face for a given input, and then learns the residual expressive component recorded in the intermediate layers of the generative model.  However, the emotion-irrelevant disturbance still persists in the intermediate layers and DeRL requires paired images for each subject. Although our method draws inspiration from DeRL, we make a different assumption for emotion extraction.

\subsection{Representation Learning for FER}
Owing to the inherent ambiguity (or uncertainty) of facial expressions, real-world FER datasets often contain mislabeled annotations, also known as noisy labels. 
Recently, much attention has been focused on learning reliable and discriminative expression features from noisy data \cite{she2021dive,zhang2022learn,zhang2021relative,chen2023static}. To this end, some prior work modifies the primary loss function or selects cleaner samples to train FER models  \cite{zhang2022learn,wang2020suppressing}.  SCN \cite{wang2020suppressing} presents a self-cure network to suppress the uncertainties in facial expression data and promote the learning of robust features. DMUE \cite{she2021dive} tackles the ambiguity problem by exploring the latent label distribution and estimating pairwise relationships between instances. RUL \cite{zhang2021relative} treats facial expressions uncertainty as a relative concept and learns it  based on the relative difficulty of two samples using feature mixup.  EAC \cite{zhang2022learn} builds an imbalanced framework and utilizes an erasing and flipping consistency loss to prevent the model from memorizing noisy samples during training.

The presence of noisy labels in FER datasets can significantly impair the performance of deep networks when conventional supervised learning is applied.
Previous work \cite{wang2020suppressing,zeng2022face2exp} has experimentally demonstrated that FER models trained solely on a specific in-the-wild FER dataset \cite{mollahosseini2017affectnet,li2017reliable,barsoum2016training} from scratch are inferior to those have already pre-trained on a large-scale upstream dataset like Ms-Celeb-1M for FR \cite{guo2016ms} and ImageNet for image classification. Further, Face2Exp \cite{zeng2022face2exp} leverages large unlabeled FR datasets to imrpove FER performance through a meta optimization framework.  In contrast, our method explores the effectiveness of MAE pre-training for enhancing facial representations in ViTs using facial expression images without ambiguous emotion labels.

\section{Our Method} \label{sec:our-method}
\subsection{Overview}
The proposed method consists of two steps.
Firstly, we pre-train plain ViTs using unlabeled facial expression images in a self-supervised manner, obtaining strong facial expression representations. These pre-trained ViTs are then fully fine-tuned on specific FER datasets to serve as decent and transparent baselines.

\begin{figure*}[!ht]
	\centerline{\includegraphics[width=0.88\linewidth]{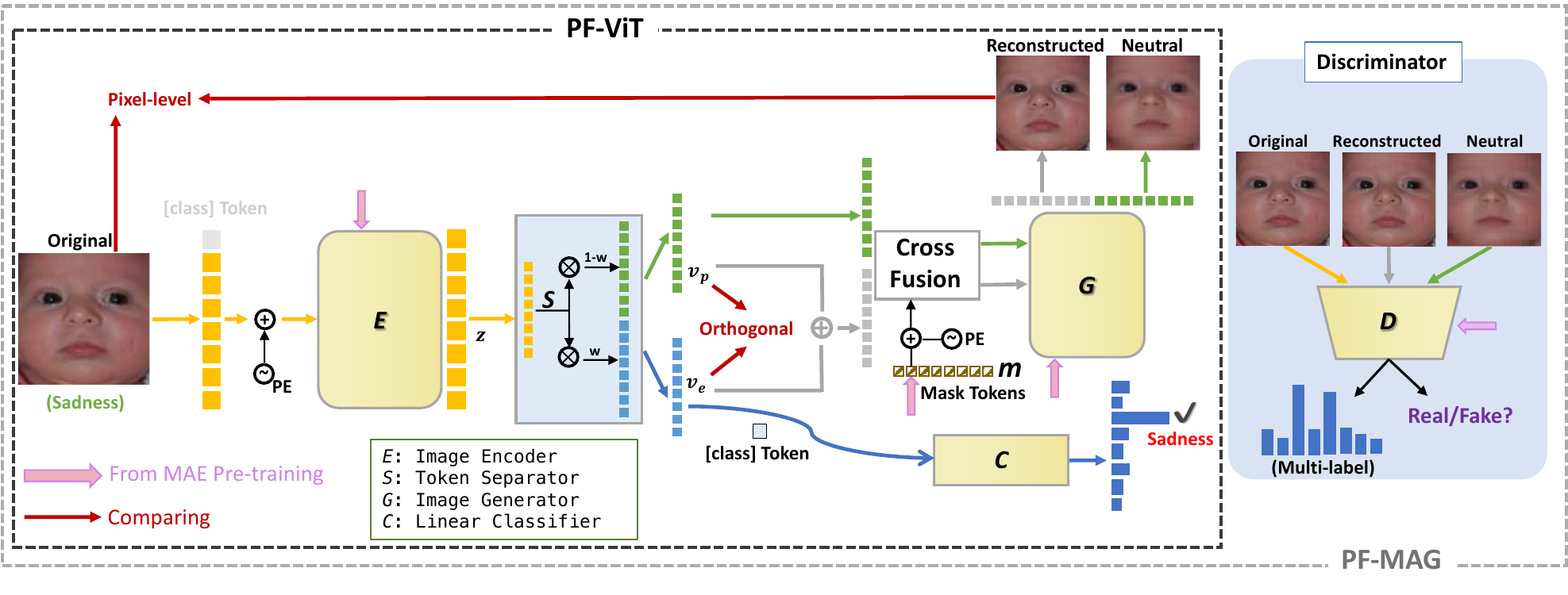}}
	   \caption{Overview of PF-ViT, which is trained in the framework of PF-MAG, a plain ViT-based GAN.  
	   Here, we reuse the pre-trained ViT encoder from our MAE pre-training stage as $E$,  and also reuse the mask tokens $\boldsymbol{m}$. Similarly, the initial $G$ and $D$ are the copies of the ViT decoder used in MAE pre-training. The indentity-independent emotion representation $\boldsymbol{v_{e}}$ and emotionless representation $\boldsymbol{v_{p}}$ are orthogonal to each other. We encourage PF-ViT to reconstruct the original face  by feeding $\left( \boldsymbol{v_{e}} + \boldsymbol{v_{p}} \right)$ to its image generator $G$, and to generate a realistic poker face preserving all the emotion-irrelevant detail of the input face when only $\boldsymbol{v_{p}}$ is used. PF-ViT and the discriminator $D$ are trained adversarially. During testing, PF-ViT classifies the facial expression based on $\boldsymbol{v_{e}}$.
	   }
	\label{fig:PFViT}
\end{figure*}

Next, we introduce the multi-task FER model, PF-ViT, which separates and classifies the emotion of the input face by generating the corresponding poker face that preserves all emotion-irrelevant details. As illustrated in Fig. \ref{fig:PFViT}, we train PF-ViT in a GAN framework without the requirement for paired images of subjects, where the pre-trained ViTs from the previous step are reused.

\subsection{Preliminary: ViT Pre-Training and Our Baselines} \label{sec:vitbaseline}
\textbf{ViT backbones}. Vision Transformers (ViTs) \cite{dosovitskiy2021an,vaswani2017attention} with the global attention mechanism have gained significant popularity as an alternative to Convolutional Neural Networks (CNNs) in various computer vision tasks. They excel at capturing long-distance relationships across the entire image. Here, we select three vanilla ViT variants, namely ViT-Base \cite{dosovitskiy2021an}, ViT-Small and ViT-Tiny, to serve as image encoders for the FER task, where ViT-Small and ViT-Tiny were originally introduced in  DeiT \cite{touvron2021training}. 

Given an input facial image $\boldsymbol{x} \in \mathbb{R}^{W \times H \times C}$ of width $W$, height $H$ and channel $C$, we reshape it into a sequence of flattened 2D image patches  $\boldsymbol{x_p} \in \mathbb{R}^{N \times \left( P^2 \cdot C \right)}$, in which the resolution of each patch is $P \times P$, and $N=H W / P^2$.  Fixed 2D sin-cos position embeddings \cite{chen2021empirical} are used to retain positional information for Transformer blocks.
The ViT encoder, denoted as $E$, maps the input to a latent representation, i.e., a sequence of visual tokens $\boldsymbol{z_p} \in \mathbb{R}^{\left(N+1 \right) \times L}$, where the first token $\boldsymbol{z_p}^{0}$ is the appended learnable {\tt [class]} token and $L$ represents the feature dimension of each latent token across its layers. 
In our method, ViT-Base (ViT-B), ViT-Small (ViT-S) and ViT-Tiny (ViT-T) all use a patch size of $P \times  P =16 \times 16$ pixels and they are only different in the embedding dimension $L$ and the number of attention heads, which are summarized in Table \ref{MAEConfig}.  

\begin{table}[!ht] 
	\caption{Configurations of the ViT encoders / (decoders) in our MAE pre-training stage.  The configurations of the decoders are shown in the brackets.
	 We follow the asymmetric encoder-decoder architecture designed in \cite{he2022masked}. For different ViT variants, we maintain a consistent embedding dimension per attention head across all layers in both encoders and decoders. Specifically, we set the dimension per head to be 64 for encoders \cite{touvron2021training} and 32 for decoders.}
	\centering
	\resizebox{1\columnwidth}{!}{
		\begin{tabular}{c|c|c|c|c}
			\toprule  
			$\begin{array}{c}\text { MAE} \\ \text {encoder}\end{array}$   &  $\begin{array}{c}\text { embedding } \\ \text { dimension }\end{array}$   &\#layers  & \#heads  &\#params  \\   
			
			\midrule    
			ViT-T & 192 / (128)  & 12 / (8)      &3 / (4)    & 5.34M /  (1.59M)   \\
			ViT-S & 384 / (256)    &12 / (8)   & 6 / (8)   & 21.29M /  (6.32M) \\
			ViT-B & 768 / (512)   &12 / (8)      &12 / (16)  & 85.05M  / (25.22M) \\

			\bottomrule
	\end{tabular}}
	\label{MAEConfig}
\end{table}

Unfortunately, ViTs typically require a larger amount of data to train compared with CNNs, as they lack strong inductive biases. The widely used FER datasets in the wild \cite{mollahosseini2017affectnet,li2017reliable,barsoum2016training} are much smaller than general image classification datasets like ImageNet. Consequently, pre-trained Transformers and even CNNs for face recognition task are utilized by prior work \cite{xue2021transfer,ma2021facial,li2021mvt} for satisfactory FER performance. It has been experimentally demonstrated that  MAE pre-training \cite{he2022masked} can effectively help ViTs learn a wide range of visual semantics and acquire excellent performance in different downstream tasks.

\textbf{MAE pre-training}.   We train ViTs as Masked Autoencoders  \cite{he2022masked} on AffectNet \cite{mollahosseini2017affectnet} dataset (the largest one for FER task publicly available up to now) without noisy and biased emotion labels, enabling us to obtain stronger representations for expressive faces.  Let $E$ denote the ViT encoder, $G$ represent the lightweight ViT decoder, and $\boldsymbol{m}$ denote the learnable mask tokens introduced in MAE structure. The corresponding configurations of decoder $G$ used in our experiments are provided in Table \ref{MAEConfig}.
\emph{For facial expressions, different emotions often involve shared facial muscle movements}. Therefore, during the pre-training stage, facial expressions with low occurrence frequencies in datasets, for example, fear, can benefit from the structure learning of more frequent facial expressions, for example, happiness. The MAE pre-training encourages the encoder to learn both local and global facial structures by reconstructing the missing pixels of randomly masked image patches, which is achieved by minimizing the loss as follows:
\begin{equation}
	\mathcal{L}_{\text {mae}}(E, G, \boldsymbol{m})=\mathbb{E}_{\boldsymbol{x}}{ \left \|G \left(E \left( \left(1 -  \mathcal{M}\right) \odot \boldsymbol{x_p}\right) \right) -  \mathcal{M}\odot \boldsymbol{x_p} \right \|_2 },
	\label{eq:MAEloss}
\end{equation}
where $\mathcal{M} \in\{0,1\}^{N}$ is the  token-level random mask that decides where to drop the image patches, and $\odot$ denotes element-wise multiplication. Thus, the mask ratio $r_{mask}$ is computed as:
\begin{equation}
	r_{mask}=\frac{\sum \mathcal{M}}{N}.
	\label{eq:r_mask}
\end{equation}
As a consequence, only a subset of the image patches from the input are visible to the lightweight MAE  decoder.
Following He et al. \cite{he2022masked}, we reconstruct the normalized pixel values of each masked image patch as the target output.

\textbf{MAE-ViT baselines for FER}. 
After completing the MAE pre-training, we append a single-layer linear classifier to the output {\tt [class]} token of each pre-trained  ViT encoder variant and proceed with fine-tuning on specific FER training sets.  Subsequently,  we evaluate the performance of these classification ViTs on FER testing sets, establishing our ViT baselines for FER. All training configurations remain consistent with those in MAE \cite{he2022masked} unless mentioned otherwise. For convenience, we refer to these fine-tuned ViTs as MAE-ViTs.

\subsection{PF-ViT: Poker Face Vision Transformer}

\textbf{Model structure.}
The framework of \textbf{P}oker \textbf{F}ace \textbf{MA}E-\textbf{G}AN (\textbf{PF-MAG}) is shown in Fig. \ref{fig:PFViT}, from which we can detach \textbf{P}oker \textbf{F}ace \textbf{V}ision \textbf{T}ransformer (\textbf{PF-ViT}) for simultaneous facial expression recognition (FER) and poker face generation.  PF-MAG is composed of PF-ViT and a composite discriminator $D$.  PF-ViT consists of four key components: 1) an image encoder $E$, which maps the input face image $\boldsymbol{x}$ to a representation $ \boldsymbol{z}$, which is a sequence of visual tokens encoding all the information about the facial image, 2) a token separator $S$, which decomposes the representation $ \boldsymbol{z}$ token-wisely into a pure emotional component $\boldsymbol{v_{e}}$  and an orthogonal residue $\boldsymbol{v_{p}}$, 3) a image generator $G$, responsible for reconstructing the original input and synthesizing the emotionless poker face using $ \boldsymbol{z}$ and mask tokens $\boldsymbol{m}$, 4) a classification head $C$, which infers the separated emotion component $\boldsymbol{v_{e}}$. The  discriminator $D$ distinguishes the facial expression image generated by $G$ and is trained adversarially with PF-ViT. During testing, both $G$ and $D$ are removable.

\emph{PF-MAG greatly benefits from the MAE pre-training} conducted in Sec. \ref{sec:vitbaseline}.
In this case, $E$ is simply a \textbf{copy} of the MAE encoder,  $D$ and $G$ are the copies of the MAE decoder. Also, we reuse the mask tokens $\boldsymbol{m}$, with each mask token representing a shared, learned vector.
Assuming $\boldsymbol{z} \in \mathbb{R}^{(N+1) \times L}$, where $L$ represents the embedding dimension of each token, and $(N+1)$ corresponds to the total number of image tokens plus a {\tt [class]} token. 
Both $E$ and $D$  utilize self-attention operations on the latent tokens. By contrast,  $G$ utilizes cross-attention operations on the decomposed latent tokens $\boldsymbol{v_{e}}$ (or $\boldsymbol{v_{p}}$) and mask tokens $\boldsymbol{m}$.
The mask tokens introduced in MAE \cite{he2022masked} indicates the presence of a missing image patch to be predicted, while \textbf{\emph{we use mask tokens $\boldsymbol{m} \in \mathbb{R}^{(N+1) \times L}$ to prompt the generator $G$ of PF-ViT to ``edit'' all the image patches of the input face}}.
Our experiments suggest that mask tokens $\boldsymbol{m}$ are crucial for $G$  to successfully reconstruct the original face and generate the poker face simultaneously. 
The emotional features can be enhanced  by comparing the synthesized poker face and original input face.

Let $(\mathcal{X}, \mathcal{Y})$ denote an in-the-wild FER dataset, in which each facial image $\boldsymbol{x} \in \mathbb{R}^{W \times H \times C}$ belongs to one of the $K$ basic expression categories, and $y \in \{1,2, \cdots, K\}$ represents the ground-truth label of $\boldsymbol{x}$. Supposing the neutral label in the current FER dataset is $c$.
The ViT encoder $E$ maps the flattened 2D patches $\boldsymbol{x_p} \in \mathbb{R}^{N \times \left( P^2 \cdot C \right)}$ reshaped from the input image $\boldsymbol{x} \in \mathbb{R}^{W \times H \times C}$ to the latent representation $\boldsymbol{z}$, briefly denoted by:
\begin{equation}
	\boldsymbol{z} = E \left( \left[\boldsymbol{x}_{\text {class }} ; \boldsymbol{x}_p^1 \boldsymbol{E} ; \boldsymbol{x}_p^2 \boldsymbol{E} ; \cdots ; \boldsymbol{x}_p^N \boldsymbol{E} \right]+\boldsymbol{E}_{pos} \right),
	\label{eq:encodermap}
\end{equation}
where $\boldsymbol{x}_{\text {class}}$ is a trainable {\tt [class]} token taken as image represent, $\mathbf{E} \in \mathbb{R}^{\left(P^2 \cdot C\right) \times D}$ represents the linear projection layer, and $\boldsymbol{E}_{pos} \in \mathbb{R}^{(N+1) \times L}$ denotes the fixed 2D sin-cos position embeddings.

The token separator S incorporates the standard squeeze and excitation block (SE) \cite{hu2018squeeze}  to generate a feature selection mask. However, we modify the activation function in the SE block from ReLU to GELU, resulting in a modified version known as Adapter  \cite{houlsby2019parameter}.
The token separator $S$ decomposes the representation  $\boldsymbol{z}$ token-wisely into an emotion-relevant component $\boldsymbol{v_{e}} \in \mathbb{R}^{(N+1) \times L}$ and an orthogonal residue $\boldsymbol{v_{p}} \in \mathbb{R}^{(N+1) \times L}$. During training, we encourage $\boldsymbol{v_{e}}$ to contain pure emotional features yet $\boldsymbol{v_{p}}$ to retain the details of the input image except for the facial muscle movements, as shown in Fig. \ref{fig:separation}. This process is simply described as follows:
\begin{equation}
	\left\{\begin{array}{l}
		\boldsymbol{v_{e}} = \boldsymbol{z} \odot  \sigma\left( S(\boldsymbol{z})  \right) \\
		\boldsymbol{v_{p}} = \boldsymbol{z} \odot  \left(1 -  \sigma\left( S(\boldsymbol{z}) \right)  \right)\\
		\text { s.t.}    \left\lvert\boldsymbol{v_{e}} \boldsymbol{v_{p}}^{\top}\right\lvert / \left( \left\|\boldsymbol{v_{e}}\right\|_2  \cdot \left\|\boldsymbol{v_{p}}\right\|_2\right) \approx 0,
	\end{array}\right.
	\label{eq:decompose}
\end{equation} 
where $\sigma$ represents the sigmoid function, $\sigma(S(\boldsymbol{z}))  \in [0,1]^{L+1}$ is the resulting element-wise selection mask for the emotional features, and we force $\boldsymbol{v_{e}}$ and $\boldsymbol{v_{p}}$ to be perpendicular in the latent space. Although the selection mask is applied independently to each token of  $\boldsymbol{z} \in \mathbb{R}^{(N+1) \times L}$, the earlier  multi-head self-attention and MLP blocks within ViT make sure each output token has already captured the global information across all other tokens.

The image generator $G$ receives the visual tokens decomposed by the separator $S$ (i.e., $\boldsymbol{v_{e}}$ and $\boldsymbol{v_{p}}$), as well as the mask tokens borrowed from MAEs, as input. It then produces two outputs: the reconstructed face image $\hat{\boldsymbol{x}}$, which retains the original emotion of the input ${\boldsymbol{x}}$, and the synthesized poker face $\tilde{\boldsymbol{x}}$, which is an emotionless version of ${\boldsymbol{x}}$.
This computation can be expressed as follows:
\begin{equation}
	\left\{\begin{array}{l}
		\hat{\boldsymbol{x}}  = G \left( MCA \left (\boldsymbol{m} + \boldsymbol{E}_{pos}, \boldsymbol{v_{e}} + \boldsymbol{v_{p}}, \boldsymbol{v_{e}} + \boldsymbol{v_{p}} \right) \right) \\
		\tilde{\boldsymbol{x}}  = G \left( MCA \left (\boldsymbol{m} + \boldsymbol{E}_{pos}, \boldsymbol{v_{p}}, \boldsymbol{v_{p}} \right ) \right ),
	\end{array}\right.
	\label{eq:comput_x}
\end{equation}
in which $MCA$ is a cross-fusion module implemented by a multi-head self-attention block ($MSA$) and mask tokens $\boldsymbol{m}$ play the role of query matrix. The cross-attention mechanism is employed to capture the relationships among the input tokens $Q^{\prime} \in \mathbb{R}^{(N+1) \times d}$, $K^{\prime} \in \mathbb{R}^{(N+1) \times d}$, $V^{\prime} \in \mathbb{R}^{(N+1) \times d}$, which is defined as:
\begin{equation}
	\operatorname{CA}(Q^{\prime}, K^{\prime}, V^{\prime})=\operatorname{Softmax}\left(Q^{\prime} {K^{\prime}}^{\top} / \sqrt{d}\right) V^{\prime},
\end{equation}
where $d$ is the embedding dimension per attention head used in the lightweight ViT generator $G$, and $Q^{\prime}$, $K^{\prime}$, $V^{\prime}$ are projected from the three input token sequences described in Eqn. \ref{eq:comput_x}, respectively. As mentioned earlier, the mask tokens $\boldsymbol{m}$ along with the position embeddings $\boldsymbol{E}_{pos}$ guide the generator $G$ to predict all the image patches with or without emotion content, according to the specific requirements. In this case, all image pathes of the input are visible to the generator $G$.

The discriminator $D$  plays a dual role in our GAN model, PF-MAG: 
\begin{itemize}
	\item [1)] It distinguishes between the synthesized facial images ($\hat{\boldsymbol{x}}$, $\tilde{\boldsymbol{x}}$) and the real facial image ($\boldsymbol{x}$),
	\item [2)] It recognizes the expressed emotions in images using a multi-label classification manner.
\end{itemize}
As illustrated in the right of Fig. \ref{fig:PFViT}, $D$ has a total of $(K+1)$ output units. The first $K$ unitis are dedicated to class-wise emotion discrimination,  with each unit representing one of the $K$ basic emotion categories.  And the last unit is responsible for discriminating between real and fake images. The proposed multi-output structure is crucial.
If we were to discard the last output unit responsible for real/fake image discrimination, \textbf{\emph{the generator $G$ could potentially exploit a ``shortcut'' to deceive the discriminator $D$}}. In such a scenario, $G$ might prioritize masking important local regions of the input face that convey emotional information, resulting in a ``masked'' facial image that appears emotionless to $D$ but does not look real.
Both $D$ and $G$ are lightweight ViTs derived from the MAE decoder pre-trained in Sec. \ref{sec:vitbaseline}. However, they are trained to fulfill different roles in the GAN model.

Finally, the single-layer linear classifier $C$ predicts the facial expression class based on the {\tt [class]} token in the separated emotion representation $\boldsymbol{v_{e}}$. The inferred emotion class is determined as follows:
\begin{equation}
	\hat{y}= \underset{k \in \{1,2, \cdots, K\}}{\operatorname{argmax}}  C\left ( \boldsymbol{v_{e}} \right )_k.
\end{equation}
During training, the classifier $C$ is supervised to output the ground-truth emotion $y$, ensuring that $\boldsymbol{v_{e}}$ contains underlying emotional factors.

\textbf{Loss functions.}
The decomposed representation $\boldsymbol{v_{e}}$ is fed into the classifier $C$,  encouraging the encoder $E$ and separator $S$ to extract and separate fine-grained emotion representations. 
The cross entropy loss is applied to supervise the $K$-emotion classification task:
\begin{equation}
	\mathcal{L}_{emo}\left(E, S, C\right)=-\mathbb{E}_{(\boldsymbol{x}, y)} \sum_{k=1}^K \mathbb I_{[y=k]} \log  \frac{e^{ C\left ( \boldsymbol{v_{e}^{\boldsymbol{x}}} \right )_{k}}}{\sum_{j=1}^K  e^{C\left ( \boldsymbol{v_{e}^{\boldsymbol{x}}} \right )_{j}}},
\end{equation}
where $\mathbb I$ is the indicator function, which equals 1 if the condition is true, and 0 otherwise.

The generator $G$ decodes the tokens processed by the cross fusion module $MCA$ into synthetic facial images, namely $\hat{\boldsymbol{x}}$ and $\tilde{\boldsymbol{x}}$, which are regarded as fake images by the discriminator $D$. 
And we define the discrimination loss as follows:
\begin{equation}
	\begin{aligned}
	 \mathcal{L}_{dis} & \left(E, S, G, D \right)  = -\mathbb{E}_{(\boldsymbol{x^{\prime}}, y)} \sum_{k=1}^{K+1} [  d^k \log \left( \sigma \left(   D(\boldsymbol{x^{\prime}}) \right) \right) \\
	& +  (1-d^k) \log \left(1 - \sigma \left(  D(\boldsymbol{x^{\prime}}) \right) \right) ],
	\end{aligned}
\end{equation}
where
$\boldsymbol{x^{\prime}} \in \{\boldsymbol{x}, \tilde{\boldsymbol{x}}, \hat{\boldsymbol{x}} \}$,  and $d^{K+1}=1$ if the input is a real image, otherwise $d^{K+1}=0$, indicating the current input is a generated image. For $1\leq k \leq K$ in the equation above, we set:
\begin{equation}
	d^k = \left\{\begin{array}{l}
		\mathbb I_{[c=k]}, \quad if \:\:\: \boldsymbol{x^{\prime}}=\tilde{\boldsymbol{x}} \\
		\mathbb I_{[y=k]}, \quad  else.
	\end{array}\right.
\end{equation}
With the supervision of loss $\mathcal{L}_{dis}$, the discriminator $D$ learns to recognize the emotion displayed in the real facial image when $\boldsymbol{x^{\prime}}=\ \boldsymbol{x}$ and further guides the generator $G$ to generate the corresponding neutral face $\tilde{\boldsymbol{x}}$ as  realistically as possible or high-quality reconstruction $\hat{\boldsymbol{x}}$ of the original input face $\boldsymbol{x}$.
During training, when $\boldsymbol{x^{\prime}} \in \{\tilde{\boldsymbol{x}}, \hat{\boldsymbol{x}} \}$, \emph{the discriminator $D$ plays an adversarial game with PF-ViT} (strictly speaking, its encoder $E$, separator $S$ and generator $G$).

To alleviate the challenge of generating facial expressions without paired images, we introduce a reconstruction loss:
\begin{equation}
	\begin{aligned}
	\mathcal{L}_{rec}\left(E, S, G \right) & =  \mathbb{E}_{\left(\boldsymbol{x}, y\right)}\left\|\hat{\boldsymbol{x}}-\boldsymbol{x}\right\|_2  \\
	 & +\mathbb{E}_{\left(\boldsymbol{x}, y\right)} \mathbb{I}\left[y=c\right]\left\|\tilde{\boldsymbol{x}} - \boldsymbol{x} \right\|_2.
	\end{aligned}
	\label{eq:reconstructloss}
\end{equation}
This loss function comprises two terms. The first term measures the pixel-level difference between the reconstructed face $\hat{\boldsymbol{x}}$ and the original input face $\boldsymbol{x}$. Meanwhile, the second term focuses on the discrepancy between the generated poker face $\tilde{\boldsymbol{x}}$ and the original input face and only applicable when the current input is a neutral expression with class label $c$.

The face reconstruction task described in Eqn. \ref{eq:reconstructloss} can provide pixel-level supervision signals, favoring the representation learning for both the encoder $E$ and generator $G$. Additionally, it serves as a constraint for the token separator $S$, ensuring that the combination of the decomposed components, i.e., $\left(\boldsymbol{v_{e}} + \boldsymbol{v_{p}}\right)$, preserves all the essential information of the input face, as described in Eqn. \ref{eq:comput_x}.

\textbf{Model training.}
During training, the image encoder $E$, token separator $S$  and FER classification head $C$ of PF-ViT are trained collaboratively to extract discriminative expression features, \textbf{minimizing} the loss $\mathcal{L}_{emo}$. FER supervision signals applied to $C$ prevent the information collapse in the separated emotional feature space.
The discriminator $D$ of PF-MAG is trained adversarially with $E$, $S$ and the generator $G$ of PF-ViT: $D$  tries to \textbf{minimize} the loss $\mathcal{L}_{dis}$ when $\boldsymbol{x^{\prime}} \in \{\boldsymbol{x}, \tilde{\boldsymbol{x}}, \hat{\boldsymbol{x}} \}$, while $E$, $S$ and $G$ make best to \textbf{maximize} it when $\boldsymbol{x^{\prime}} \in \{\tilde{\boldsymbol{x}}, \hat{\boldsymbol{x}} \}$.
Meanwhile, the orthogonal constraint in Eqn. \ref{eq:decompose} aids in the separating of $\boldsymbol{v_{e}}$ and $\boldsymbol{v_{p}}$.
In order to fool the discriminator $D$, the synthesized poker face $\tilde{\boldsymbol{x}}$ needs to resemble a true neutral face, and the reconstructed face $\hat{\boldsymbol{x}}$ should look nearly the same as the original input $\boldsymbol{x}$ (refer to Eqn. \ref{eq:reconstructloss}). 

As a result, the representation $\boldsymbol{v_{p}}$ retains all the facial details except for emotion, whereas the representation $\boldsymbol{v_{e}}$ only contains the emotional information.
In other words, the representation space of $\boldsymbol{v_{e}}$ primarily emphasizes and extracts the emotional component. The inclusion of the auxiliary task of image generation compels our FER model to capture more facial and emotional details.

To the best of our knowledge, training vanilla ViT-based GAN for image generation is quite challenging, even with sophisticated techniques  \cite{leevitgan}. Fortunately, \emph{PF-ViT with the face generation task can be easily trained by leveraging our MAE representations} (including the mask token) learned from a large number of facial expression images without labels.

\begin{figure*}[h]
	\centerline{\includegraphics[width=0.9\linewidth]{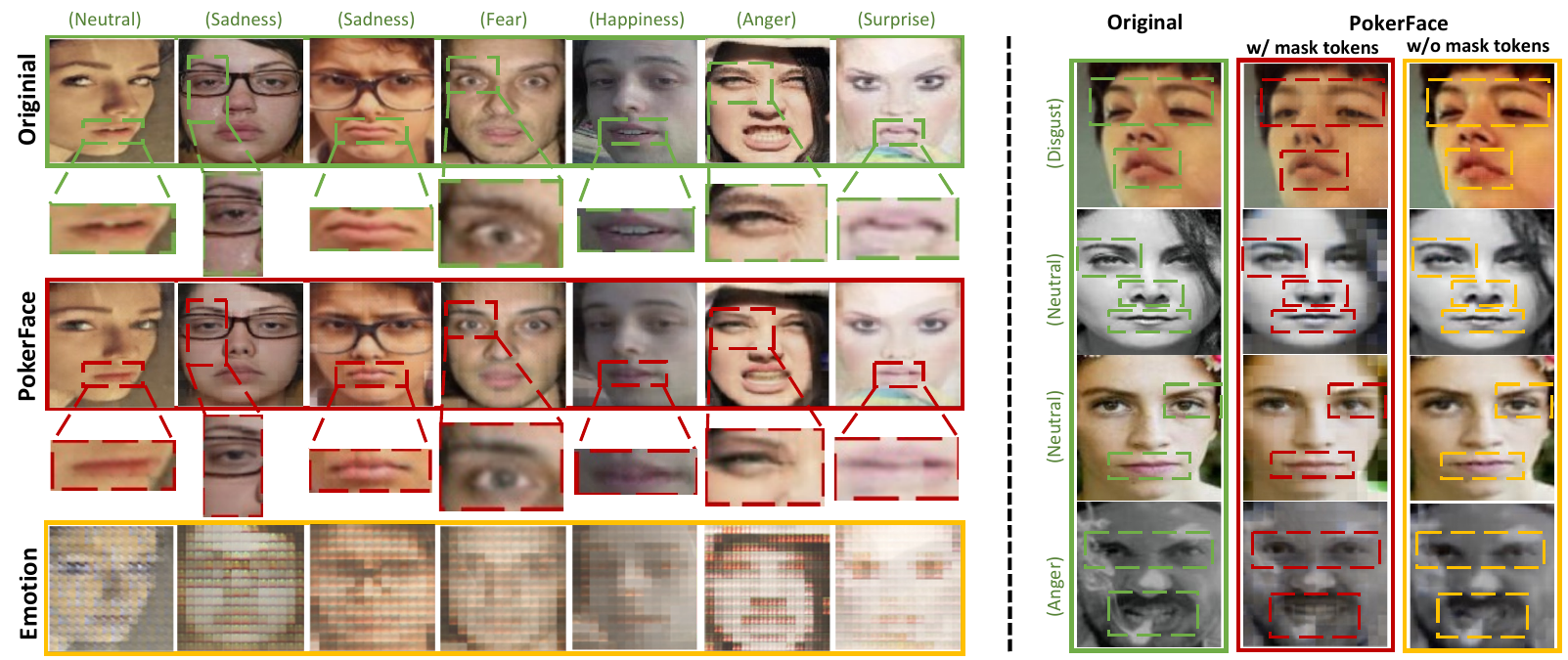}} 
	   \caption{Visualization of the emotion separation and poker face generation. The PokerFace and Emotion images are restored by the generator $G$ of PF-ViT from the disentangled components $\boldsymbol{v_{e}}$ and $\boldsymbol{v_{p}}$, respectively.
		\textbf{Left:} both visual tokens and mask tokens are fed into the generator $G$. By comparing the original expressive face and synthesized poker face, we can see the discrepancy in terms of emotional details. Note that we train PF-ViT only using unpaired images.
		\textbf{Right:} apart from the visual tokens produced by the token separator $S$, mask tokens are also crucial for the generator $G$ to synthesize satisfactory poker faces. 
	   }
	\label{fig:pokerface}
\end{figure*}

\section{Experiments} \label{sec:our-experiment}
\subsection{Datasets}
\textbf{RAF-DB} \cite{li2017reliable} is a widely used in-the-wild FER dataset that contains 29,672 facial images collected from the Internet. Each sample in the dataset is independently labeled by around 40 annotators. RAF-DB provides labels for seven basic expressions and eleven compound expressions. In our experiments, we focus on the seven basic expressions, namely neutral, happiness, surprise, sadness, anger, disgust, and fear, involving 12,271 images for training and 3,068 images for testing.

\textbf{AffectNet} \cite{mollahosseini2017affectnet} is currently the largest real-world FER dataset publicly available containing over 1 million (1M) facial images collected from the Internet. It holds manually and automatically labeled samples. And our experiments use the manually labeled images, following the convention in prior work  \cite{xue2021transfer,ma2021facial,li2021mvt,wang2020suppressing,zhang2021relative}. There are two different choices for result comparison denoted by \textbf{AffectNet-7} and \textbf{AffectNet-8}, considering the said seven basic expressions and the seven basic expressions plus contempt, respectively. AffectNet-7 includes 283,901 images for training and 3,500 images for testing, while AffectNet-8 contains 287,568 images for training and 4,000 images for testing. 

\textbf{FERPlus} \cite{barsoum2016training} is an extended version of FER2013, including 28,709 training, 3,589 validation and 3,589 testing grayscale images, collected using the Google search engine. Each image in the dataset, with the spatial extent of 48$\times$48 pixels, is labeled by 10 annotators to one of the eight emotion classes (the seven basic expressions plus contempt). The validation set is commonly used for model training as well.

\subsection{Implementation Details}
Following the conventional practice, we perform in-the-wild FER in two steps: (1) the face is detected and aligned, and further resized to 224$\times$224 pixels, (2) the facial expression is classified.
We implement our method using Python and PyTorch, and use automatic mixed precision (Amp) supported by Nvidia \emph{Apex} PyTorch library to seep up the experiments.

\textbf{Building FER baselines based on ViTs}.  
AffectNet dataset contains 400,000 manually labeled images. We clean the dataset based on the label file and perform face alignment using OpenFace \cite{baltrusaitis2018openface}. This results in 269,316 facial images for our self-supervised MAE pre-training, dubbed \textbf{AffectNet-270K} dtaset. 
Each variant of ViT (ViT-Tiny, ViT-Small and ViT-Base) is pre-trained as the MAE encoder for 300 epochs. Subsequently, the pre-trained ViTs are fine-tuned on the downstream FER training sets \cite{mollahosseini2017affectnet,li2017reliable,barsoum2016training} respectively, using the Adam optimizer with a batch size of 32 and an initial learing rate of 1e-4. During training, we reduce the learning rate by a factor of 10 if no improvement is observed for 30 epochs. Every training process is finished within 100 epochs.  Common data augmentation techniques, such as random scaling and flipping, are employed unless mentioned otherwise.
In addition, we train and evaluate the ViT variants initialized with the regular ImageNet1K weights\footnote {Strictly speaking, they are pre-trained on ImageNet21K dataset and fine-tuned on ImageNet1K dataset, which are publicly available in \emph{timm} PyTorch library.}  provided by Steiner et al. \cite{steiner2021train,dosovitskiy2021an} for the FER task, obtaining more baseline results.

\textbf{Implementation details of PF-MAG}.  
According to our experimental findings, directly applying the orthogonal constraint in Eqn. \ref{eq:decompose} to $\boldsymbol{v_{e}}$ and $\boldsymbol{v_{p}}$ immediately after the token separator $S$ fails to  yield satisfactory disentanglement. To address this issue, we leverage the pre-trained Incept-ResV1, a face recognition (FR) model introduced by Schroff et al. \cite{schroff2015facenet}, to extract the facial features of the images generated by $G$ based on $\boldsymbol{v_{e}}$ and $\boldsymbol{v_{p}}$. Then, these extracted features are flattened and used to comput a cosine similarity, which indirectly measures the distance between $\boldsymbol{v_{e}}$ and $\boldsymbol{v_{p}}$ in the off-the-shelf latent space of the well-trained FR model. The pre-trained FR model is fixed and incorporated into the orthogonal constraint loss during training.  \label{sec:orthogonalfacenet} 
It's important to note that we reuse the pre-trained MAE encoders and decoders, as described in Table \ref{MAEConfig}, to construct our GAN model, PF-MAG.

The encoder $E$, separator $S$, generator $G$, and classifier $C$ are optimized using the Adam optimizer with an initial learing rate of 2e-5. The discriminator $D$ is optimized using the Adam optimizer with $\beta_1=0.5$, $\beta_2=0.999$, and an initial learing learning rate of 1e-4. We approximately balance the different loss terms according to their magnitudes. The remaining training configurations follows our ViT baselines described earlier.
It is worth noting that we did not employ grid search to subtly optimize the hyperparameters for each dataset. Nevertheless, we have achieved decent accuracy performance.

\begin{table}
	\caption{Comparison between MAE representation and conventional SL representation for FER based on \textbf{ViT-B}. Accuracy (\%) numbers are evaluated on the RAF-DB, AffectNet-8 and FERPlus testing sets. The entry marked with ``$^{*}$'' denotes the model is trained without data augmentation techniques.}\smallskip 
	\centering
	\resizebox{1\columnwidth}{!}{
		\smallskip\begin{tabular}{c|c|c|c|c}  
			\toprule
			method & pre-train data & RAF-DB  & AffectNet-8 &FERPlus \\   
			
			\midrule    
			\rowcolor{lightgray}SL$^{*}$ & ImageNet1K          & 85.14 & 57.10  &85.78 \\
			MAE$^{*}$                 & ImageNet1K          & 88.49 \small{({+3.35})} & 60.87 \small{({+3.77})}  &88.29 \small{({+2.51})} \\
			\rowcolor{cyan!20} MAE$^{*}$                 & AffectNet-270K      & \textbf{90.22} \small{({+5.08})} & \textbf{61.73} \small{({+4.63})} &\textbf{89.00} \small{({+3.22})} \\ 
			\textcolor{lightgray}{MAE}        & \textcolor{lightgray}{AffectNet-270K}     & \textcolor{lightgray}{\textbf{91.07} \small{({+5.93})}}   & \textcolor{lightgray}{\textbf{62.42} \small{({+5.32})}}  & \textcolor{lightgray}{\textbf{90.18} \small{({+4.40})}} \\
			\bottomrule
	\end{tabular}}
	\begin{tablenotes}
		\item $\dagger$ ``SL'' represents the common supervised learning method for pre-training, while ``MAE'' stands for the self-supervised MAE pre-training. The models above undergo initial pre-training and subsequent fine-tuning for the task of FER.
	\end{tablenotes}
	\label{MAEcompare}
\end{table}

\subsection{Ablation Studies}
\textbf{Competitive ViT baselines with MAE representations}.
We show the FER fine-tuning results of the pre-trained ViTs in Table \ref{MAEcompare}, whose initial representations are obtained by conventional supervised learning (SL) or self-supervised MAE learning(MAE) \cite{he2022masked}. \textbf{\emph{The MAE representation brings about remarkable and consistent accuracy improvements}} to ViT-B on the RAF-DB (+3.35\%), AfectNet-8 (+3.77\%) and FERPlus (+2.51\%) testing sets, compared to the widely used SL representations learned on ImageNet1K dataset (the second row vs. the first row). 
The FER model based on the ViT-B variant pre-trained on AffectNet-270K dataset using the MAE method achieves respectable accuracy numbers: 90.22\%, 61.73\% and 89.00\% on the RAF-DB, AffectNet-8 and FERPlus testing sets, respectively, without data augmentation.

\begin{figure*}
	\centerline{\includegraphics[width=1\linewidth]{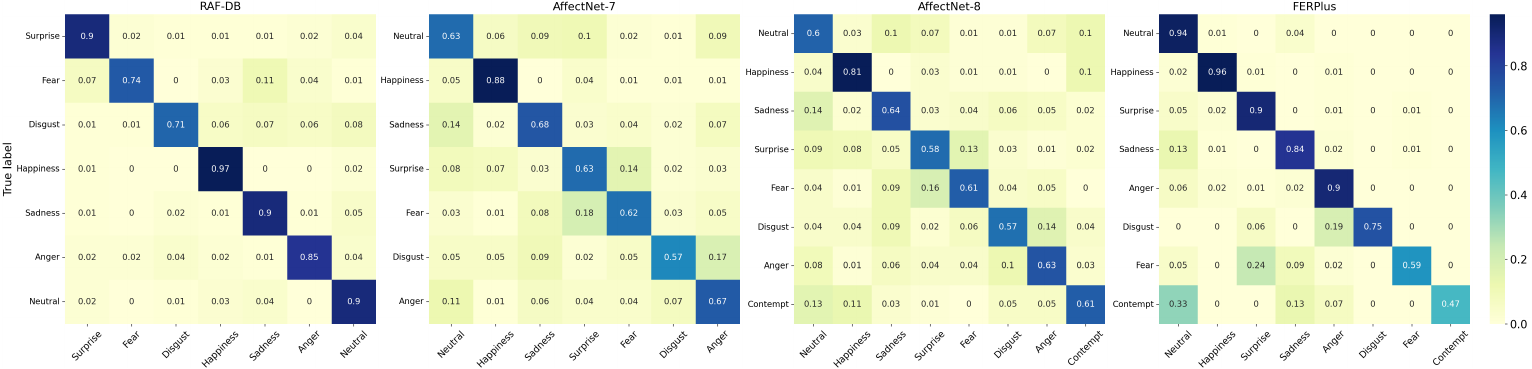}}
	   \caption{Confusion matrices of our PF-ViT on the RAF-DB, AffectNet-7, AffectNet-8 and FERPlus testing sets.  The average recognition rates across expression classes are 92.07\%, 67.23\%, 64.10\% and 91.16\% respectively.
	   }
	\label{fig:confution_matrix} 
\end{figure*}

Based on the comparison between the second and third entries in Table \ref{MAEcompare}, it can be concluded that the ViT-B FER model pre-trained with MAE self-supervised learning on a facial dataset (AffectNet-270K) significantly outperforms the one pre-trained with MAE self-supervised learning on a general image recognition dataset (ImageNet1K). It is worth noting that the AffectNet-270K dataset we utilize only contains about 1/5 of the images compared to the ImageNet1K training set.
This suggests that \textbf{\emph{a more relevant representation, which includes facial knowledge, is superior to a more general image representation}}.
Furthermore, the accuracy numbers of the ViT-B model increase to 91.07\% on RAF-DB, 62.42\% on AfectNet-8, and 90.18\% on FERPlus testing sets, respectively, when data augmentation is employed. This observation aligns with the fact that Transformers are known to benefit from larger amounts of data, indicating their ``data hunger".

\begin{table} 
	\caption{Accuracy (\%) evaluation of our FER baselines based on ViT-B, ViT-S and ViT-T on the RAF-DB, AffectNet-8 and FERPlus testing sets. MAE representations have nothing to do with the noisy and biased emotion labels. The following results are obtained by us under fair conditions.}
	\centering
	\resizebox{1\columnwidth}{!}{
		\begin{tabular}{c|c|c|c|c|c}
			\toprule  
			method & pre-train data &backbone & RAF-DB  & AffectNet-8 &FERPlus \\   
			
			\midrule   
			\multicolumn{6}{l}{\textbf{SL-ViT baselines for FER}} \\
			\cmidrule(r){1-6}  
			SL & ImageNet1K     &ViT-T      &87.03   & 57.91   &88.46  \\
			SL & ImageNet1K    &ViT-S      & 87.19  & 57.99   &88.56  \\
			\rowcolor{lightgray}SL & ImageNet1K     &ViT-B      & 87.22  &58.28    &88.91  \\
			\cmidrule(r){1-6} 
			\multicolumn{6}{l}{\textbf{The proposed MAE-ViT baselines for FER}} \\
			\cmidrule(r){1-6}  
			MAE        & AffectNet-270K    &ViT-T    & 88.72  &61.45  &88.67   \\ 
			MAE        & AffectNet-270K    &ViT-S    & 90.03  &62.06  &89.35   \\
			\rowcolor{cyan!20} MAE        & AffectNet-270K    &ViT-B    & \textbf{91.07}  & \textbf{62.42} & \textbf{90.18}  \\ 

			\bottomrule 
	\end{tabular}}
	\label{ViTbaseline}
\end{table}

\begin{table}
	\caption{Accuracy (\%) results of our PF-ViT models with different ViT variants on the RAF-DB, AffectNet-8 and FERPlus testing sets. The corresponding MAE decoders summarized in Table \ref{MAEConfig} are reused as the generator $G$ and discriminator $D$ in our GAN model, illustrated in Fig. \ref{fig:PFViT}.
	And the pre-trained MAE encoders are also reused.
	}\smallskip 
	\centering
	\resizebox{0.85\columnwidth}{!}{
		\smallskip\begin{tabular}{c|c|c|c|c} 
			\toprule  
			method & encoder $E$ & RAF-DB  & AffectNet-8 &FERPlus \\   
			
			\midrule
			\textcolor{lightgray}{MAE-ViT}    & \textcolor{lightgray}{ViT-B}   & \textcolor{lightgray}{{91.07}}   & \textcolor{lightgray}{{62.42}}  & \textcolor{lightgray}{{90.18}} \\    
			PF-ViT & ViT-T           &86.26 &58.95  &88.81 \\
			PF-ViT      & ViT-S        &87.47  &59.86   &88.94 \\
			\rowcolor{cyan!20}PF-ViT      & ViT-B      & \textbf{92.07} & \textbf{64.10}  &\textbf{91.16}  \\ 
			
			\bottomrule 
	\end{tabular}}
	\label{PFViTvariants}
\end{table}

Benefiting from the strong facial expression representations obtained through MAE pre-training using AffectNet-270K images without noisy emotion labels, vanilla ViTs are able to achieve decent FER performance, which we refer to as MAE-ViTs for simplicity. Several MAE-ViT and SL-ViT baselines based on different ViT variants (ViT-B, ViT-S and ViT-T) are provided in Table \ref{ViTbaseline}. The experiments consistently demonstrate that ``AffectNet-270K-MAE'' representations outperforms the commonly used ``ImageNet1K-SL'' representations  across different ViT backbones. Even with data augmentation techniques, the ``AffectNet-270K-MAE'' representation still leads to significant accuracy improvements of +3.85\% on the RAF-DB, +4.14\%  on the AffectNet-8, and +1.27\% on the FERPlus testing tests when ViT-B is utilized (comparing the 6th row and the 3rd row of Table \ref{ViTbaseline}).

\begin{figure}[!h]
	\centerline{\includegraphics[width=0.55\linewidth]{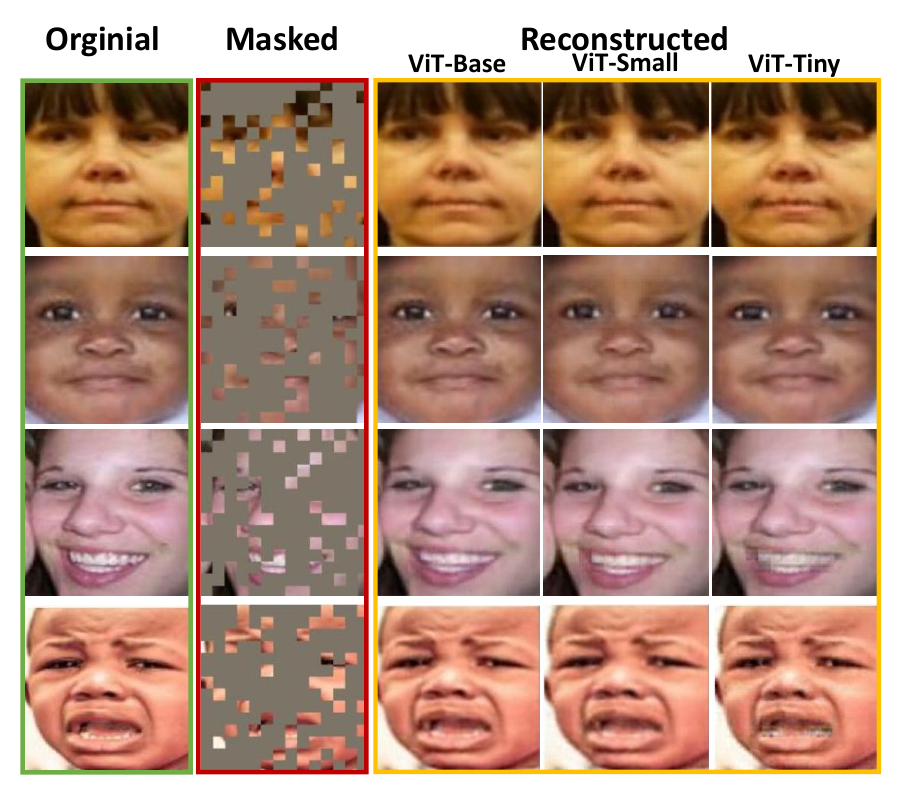}}
	   \caption{Reconstructions of maksed facial imags from RAF-DB dataset generated by MAEs using different ViT encoders, with a \textit{mask ratio} of 0.75 (i.e., 75\% of the original image is randomly masked).}
	\label{fig:MAEreconstruction}
\end{figure}

Incidentally, in Fig. \ref{fig:MAEreconstruction}, we show some reconstruction examples of the maksed facial images produced by MAE models based on ViT encoders and decoders of different scales. These MAE models are trained from scratch using AffectNet-270K dataset. It can be seen that MAE models are very good at reconstructing the missing pixels of the masked patches of the original facial image. The details of the MAE encoders and decoders configurations are presented in Table \ref{MAEConfig}.

\textbf{Component analysis of PF-ViT}.
We show the evaluation results of the proposed PF-ViT based on ViT-B, ViT-S and ViT-T in Table \ref{PFViTvariants}, leveraging our MAE representations learned from AffectNet-270K data.  In contrast to the fine-turning results of SL-ViT baselines in Table \ref{ViTbaseline}, large accuracy gaps exist among the PF-ViT models with the three ViT variants as the image encoder $E$. This indicates that \textbf{\emph{a moderately-sized ViT model with adequate  capacity is more suitable for PF-ViT to effectively perform poker face generation and emotion classification simultaneously}}. The inclusion of the auxiliary task of poker face generation further improves the performance of our PF-ViT model, compared with MAE-ViT.
Comparing the first and last rows in Table \ref{PFViTvariants}, PF-ViT outperforms MAE-ViT by 1\%, 1.68\% and 0.98\% in accuracy on the RAF-DB, AffectNet-8 and FERPlus testing sets, correspondingly.
The confusion matrices of PF-ViT are computed and presented in Fig. \ref{fig:confution_matrix}. 
Additionally, in Appendix \textcolor{red}{A} of our supplementary material, we show several typical examples of misclassifications made  by PF-ViT. These errors primarily arise from the ambiguity present in facial expressions.

To help understand how PF-ViT works, we visualize the emotional component (denoted as ``Emotion'') and emotionless poker face (denoted as ``PokerFace'') in the \textbf{Left} of Fig. \ref{fig:pokerface}. 
This visualization is achieved by separately feeding the emotional component $\boldsymbol{v_{e}}$ and emotionless component $\boldsymbol{v_{p}}$ into the generator $G$ of PF-ViT.
By comparing the Original and PokerFace images, it can be seen that the generated poker faces retain the emotion-irrelevant factors of the original input images while the emotional details are almost eliminated. 
Thanks to the orthogonal constraint applied to $\boldsymbol{v_{e}}$ and $\boldsymbol{v_{p}}$, described in Eqn. \ref{eq:decompose}, the visualized Emotion images exhibit significant differences compared to the Original and PokerFace images, even though some facial residues are sitll restored by the generator  $G$ from $\boldsymbol{v_{e}}$. 
Incidentally, the reconstruction loss defined by Eqn. \ref{eq:reconstructloss} helps stableize the the adversarial training of poker face generation, for it provides pixel-level supervision signals from paired images.

Just for reference, the poker faces generated by our PF-ViT, based on ViT-B, obtain an FID  score (Fréchet Inception Distance) \cite{heusel2017gans} of 100.38 when compared to the real faces on RAF-DB dataset. 
However, the FID score may not reliably indicate the quality of the generated facial images. \emph{In cases where the de-expression process totally fails, the FID score may improve significantly}. This is because the generated poker face would look nearly the same as the original input, leading to a much smaller distribution distance between the generated and real images.

\textbf{\emph{Enforcing strict orthogonality between the decomposed components $\boldsymbol{v_{e}}$ and $\boldsymbol{v_{p}}$ in the high-dimensional space}} immediately after the token separator $S$ \textbf{is challenging}, even though the cosine similarity (absolute value) between them significantly reduces during training. However, the poker faces synthesized by the generator $G$ still retain noticeable emotional details in this scenario. 
To evaluate the effectiveness of the FR model (Incept-ResV1) in measuring feature distance, we conducted ablation experiments by either replacing it with a image classification ResNet pre-trained on ImageNet1K dataset or removing it entirely. The results are presented in Table \ref{ablationSeparator}. 
According to the first row of the results, applying the orthogonal constraint to $\boldsymbol{v_{e}}$ and $\boldsymbol{v_{p}}$ directly leads to a significant decrease in accuracy on the RAF-DB and AffectNet-8 testing sets. 
Referring to the last entry in Table \ref{ablationSeparator}, utilizing the pre-trained FR model to firstly extract the high-level facial features from the images derived from $\boldsymbol{v_{e}}$ and $\boldsymbol{v_{p}}$ before applying the orthogonal constraint significantly improves the accuracy by large margins, compared to other ResNet models used for general image feature extraction.

\begin{table}
	\caption{Comparison of different facial feature extractor before applying orthogonal constraint on the RAF-DB and AffectNet-8 testing sets. We use the feature embeddings befor the classification heads of the  pre-trained ResNet models. Only the first entry computes the similarity between the decomposed components directly in the representation sapce immediately after the token separator $S$.}
	\centering
	\resizebox{1\columnwidth}{!}{
		\smallskip\begin{tabular}{cccccc}
			\toprule
			 $\begin{array}{c}\text {direct} \\ \text {constraint}\end{array}$  & ResNet-18 & ResNet-50  &    $\begin{array}{c}\text {Incept-ResV1} \\ \text {(FR model)}\end{array}$  & RAF-DB & AffectNet-8 \\
			\midrule
			$\checkmark$ &           &      &       &86.67 &59.71 \\
			$\checkmark$ & $\checkmark$ &   &        & 89.99 &61.68 \\
			$\checkmark$ & &$\checkmark$   &        & 90.22 & 62.71 \\
			$\checkmark$ &   &  & $\checkmark$ & 92.07 & 64.10 \\
			\bottomrule
	\end{tabular}}
	\label{ablationSeparator}
\end{table}

Further, \textbf{\emph{to demonstrate the necessity of token separation in our method}}, we remove the token separator $S$ and conduct the ablation experiments as outlined in Table \ref{ablationgan}, in which only original and reconstructed images are utilized for training. Consequently, PF-ViT degrades to the original Masked Autoencoder (MAE) \cite{he2022masked}. Here, we train and evaluate two variant GAN frameworks: 1) the MAE model trained adversarially with a discriminator $D^{'}$, which only identifies real and fake images (presented in the second row); 2) the MAE model trained adversarially with a discriminator $D^{''}$, which not only distinguishes between real and fake images but also verifies whether the facial expression of the reconstructed image is consistent with the original face (presented in the third row). The results of these two variants exhibit noticeable decreases in accuracy, with reductions of more than 1.4\%, 3.5\%, and 1.5\% accuracy on the RAF-DB, AffectNet-8, and FERPlus testing sets, respectively. In addition,  $D^{''}$ enforces emotional consistency during the reconstruction of facial images in training, and results in a slightly higher FER accuracy compared to $D^{'}$. Overall, the ablation experiments in Table \ref{ablationgan} indicate that direct adversarial training with the discriminator ($D^{'}$ or $D^{''}$) is not beneficial for the final FER model to capture nuanced expressive features but rather introduces negative factors.

\begin{table}
	\caption{Ablation study of token separation. In the second and third rows, we remove the token separator $S$, initially depicted in Fig. \ref{fig:PFViT}, and retrain the whole framework using original and reconstructed images only. The results in the second and third rows correspond to the GAN framework employing the single-task discriminator $D^{'}$ and the multi-task discriminator $D^{''}$, respectively. Accuracy (\%) numbers are evaluated on the RAF-DB, AffectNet-8, and FERPlus testing sets. Our ViT-B pre-trained on AffectNet-270K is the default encoder.}
	\centering
	\resizebox{0.9\columnwidth}{!}{
		\smallskip\begin{tabular}{cccc}
	\hline GAN structure & RAF-DB & AffectNet-8 & FERPlus \\
	\hline \textcolor{lightgray}{(PF-ViT, $D$)} & \textcolor{lightgray}{92.07} & \textcolor{lightgray}{64.10} & \textcolor{lightgray}{91.16} \\
	(MAE, $D^{'}$) & 90.35 (-1.72) & 60.07 (-4.03) & 89.15 (-2.01) \\
	(MAE,  $D^{''}$) & 90.64 (-1.43) & 60.42 (-3.68) & 89.59 (-1.57) \\
	\hline
	\end{tabular}}
	\label{ablationgan}
\end{table}

\begin{figure}
	\centerline{\includegraphics[width=0.75\linewidth]{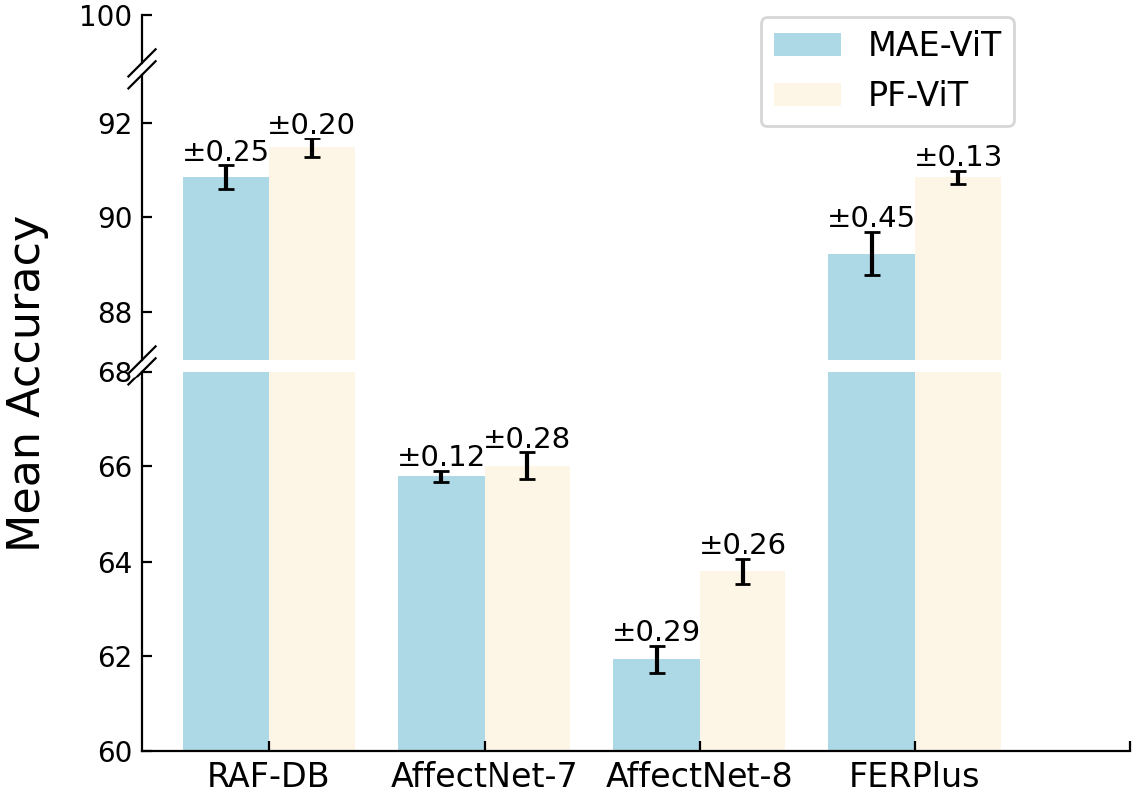}}
		\caption{Histgrams depicting the average and standard deviation of accuracy performance per dataset. In this analysis, both MAE-ViT and PF-ViT models utilize ViT-B as the image encoder. The standard deviation values are indicated on top of each histogram.
		}
	\label{fig:std}
\end{figure}

Similar to the mask tokens introduced in MAE \cite{he2022masked}, our mask tokens $\boldsymbol{m}$ along with the position embeddings $\boldsymbol{E}_{pos}$ indicate to the generator $G$ where to ``edit'' the image patches when only the emotion-irrelevant token sequence $\boldsymbol{v_{p}}$ is provided. Interestingly, \textbf{\emph{the generator $G$ cannot remove emotional contents from facial images without utilizing the mask tokens $\boldsymbol{m}$}}.  In this case, the $MCA$ module, described in Eqn. \ref{eq:comput_x}, degrades to a $MSA$ module. Then, we retrained our GAN model, but the adversarial training did not converge, and feature disentanglement failed. As a result, both  $\boldsymbol{v_{e}}$ and  $\boldsymbol{v_{p}}$ retained the input facial contents. The resulting examples generated from $\boldsymbol{v_{p}}$  are shown in the \textbf{Right} of Fig. \ref{fig:pokerface}.

\begin{table*}[!ht]
	\caption{Comparing our FER models with the existing SOTA FER models on the RAF-DB, AffectNet-7, AffectNet-8 and FERPlus testing sets. We present the highest reported FER accuracy numbers (\%) from their respective papers. The following Hybrid ViT models utilize a ResNet pre-trained on FR datasets to extract facial features befor applying Transformer blocks.
	}\smallskip
	\label{Networkcompare}
	\centering
	\resizebox{0.82\linewidth}{!}{
		\smallskip\begin{tabular}{c|c|c|c|c|c|c|c}  
			\toprule 
			method &venue &backbone &pre-training data  & RAF-DB  & AffectNet-7  & AffectNet-8  & FERPlus  \\   
			
			\midrule    
			\multicolumn{8}{c}{\textbf{Plain CNN models}} \\
			\cmidrule(r){1-8}  
		
			SCN \cite{wang2020suppressing}   &CVPR'20   &ResNet-18    &Ms-Celeb-1M  & 88.14    &-   & 60.23    &89.35    \\
			RUL \cite{zhang2021relative}   &NeurIPS'21    &ResNet-18   &MS-Celeb-1M      & 88.98    &-   &-  &-    \\
			EAC \cite{zhang2022learn}   &ECCV'22 	&ResNet-18	&MS-Celeb-1M	 &89.99    &65.32    &-   &89.64    \\
			DMUE \cite{she2021dive}   &CVPR'21 	&ResNet-18	&Ms-Celeb-1M	 &88.76    &-    &62.84   &88.64    \\
			DMUE \cite{she2021dive}   &CVPR'21 	&Res-50IBN	&-	 &89.42    &-    &63.11   &89.51    \\
			IPD-FER \cite{jiang2022disentangling}  &TAFFC'22 &ResNet-18 &CASIA-WebFace \cite{yi2014learning}  &88.89   &62.23     &-     &88.42   \\
			AMP-Net \cite{liu2022adaptive}  &TCSVT'22 &ResNet-34 &VGGFace2  &89.25   &64.54     &61.74     &-   \\
			LDLVA \cite{le2023uncertainty}   &WACV'23  &ResNet-50 &ImageNet1K  &90.51  &66.23 &- &- \\
			\cmidrule(r){1-8} 
			\multicolumn{8}{c}{\textbf{Hybrid Visoin Transformer models} (rely on GCN- or CNN-based backbones)} \\
			\cmidrule(r){1-8}  
			VTFF \cite{ma2021facial}   &TAFFC'23 &Hybrid ViT  &MS-Celeb-1M  &88.14   &-     &61.58     &88.81   \\
			TransFER \cite{xue2021transfer}  &ICCV'21  &Hybrid ViT &Ms-Celeb-1M &90.91   &66.23      &-    &90.83   \\
			MRAN \cite{chen2023multi}   &TCSVT'23  &Hybrid ViT &MS-Celeb-1M &90.03   &66.31   &62.48  &89.59   \\
			FG-AGR \cite{li2023fg}   &TCSVT'23  &Hybrid ViT &MS-Celeb-1M &90.81   &64.91   &60.69  &91.09   \\
			\cmidrule(r){1-8} 
			\multicolumn{8}{c}{\textbf{Plain Visoin Transformer models} (pure Transformers without other backbones)} \\
			\cmidrule(r){1-8}  
			MVT \cite{li2021mvt}   &Arxiv'21 	&DeiT-S \cite{touvron2021training} &ImageNet1K	 & 88.62   &64.57    & 61.40  &89.22    \\
			MTAC \cite{liu2024mtac}   &TMM'24 	&Swin-S \cite{liu2021swin} &ImageNet1K	 & 90.52   &--    & 62.28  &90.4   \\
			\cmidrule(r){1-8}
			MAE-ViT (Ours)    &TCSS'24   &ViT-T  &AffectNet-270K        & 88.72   &64.25  &61.45  &88.67  \\
			MAE-ViT (Ours)   &TCSS'24    &ViT-S   &AffectNet-270K        & 90.03   &65.53  &62.06  &89.35  \\
			\rowcolor{lightgray} MAE-ViT (Ours)   &TCSS'24     &ViT-B  &AffectNet-270K       & 91.07    &66.09   &62.42   &90.18   \\
			\rowcolor{cyan!25} \textbf{PF-ViT (Ours)}    &TCSS'24\footnotemark    &ViT-B  &AffectNet-270K            & \textbf{92.07}    &\textbf{67.23}    & \textbf{64.10}    &\textbf{91.16}  \\

			
			\bottomrule 
	\end{tabular}}
\end{table*}

We firmly believe that \textbf{\emph{all image patches of the input face are crucial for accurate and fine-grained facial expression recognition}}. To validate this belief, we present a multi-task MAE (Masked Autoencoder) model for simultaneous poker face synthesis and emotion classification, as detailed in Appendix \textcolor{red}{B} of our supplementary material. During training, the multi-task MAE only observes a subset of the image tokens, resulting in inferior performance compared to PF-ViT. This further reinforces the effectiveness and superiority of the PF-ViT model in capturing all facial information and achieving superior performance in both poker face synthesis and emotion classification tasks.

In our experiments, we have observed \textbf{\emph{fluctuations in FER accuracy}}. The varying accuracy can be attributed to several factors: 1) the current in-the-wild FER datasets, such as AffectNet \cite{mollahosseini2017affectnet} and RAF-DB \cite{li2017reliable}, are significantly class-imbalanced, 2)  noisy emotion labels are prevalent in these datasets, introducing additional challenges for accurate classification, and 3) the limited size of the testing set can contribute to variations in accuracy. In Fig. \ref{fig:std}, we present the mean and standard deviation of the overall accuracy across 3 additional runs per dataset and per model, which are trained for the same epochs as the corresponding entries listed in Table \ref{PFViTvariants}.
Here, we provide cross-dataset evaluations following \cite{chen2023static}. The PF-ViT model trained on AffectNet-8 achieves 73\% accuracy on RAF-DB, while the model trained on RAF-DB drops to 47.8\% when tested on AffectNet-8.

\footnotetext{The experimental results were first presented in our earlier preprint: \url{https://arxiv.org/abs/2207.11081v3}.}

\begin{figure}[!h]
	\centerline{\includegraphics[width=0.9\linewidth]{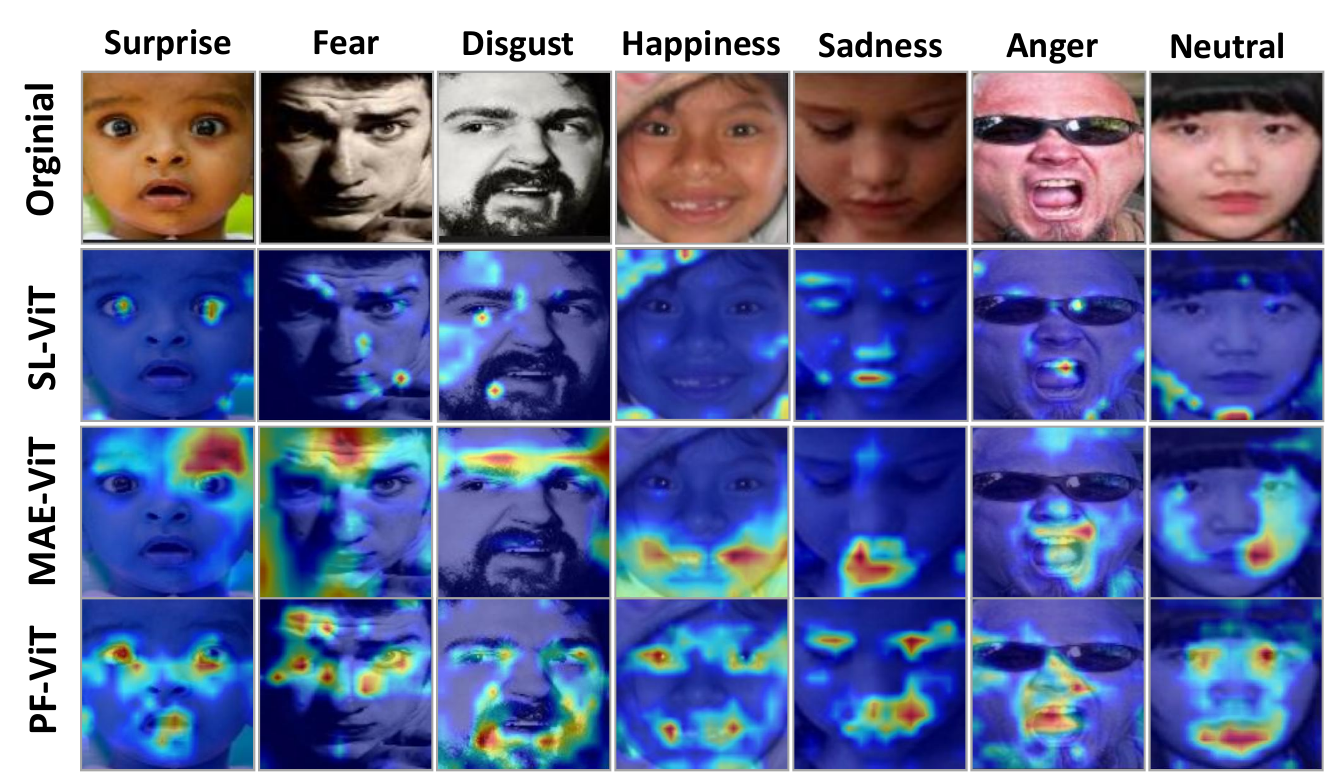}}
	   \caption{The attention maps of different facial expressions. The FER models above are all based on ViT-B  to ensure a fair comparison. With the inclusion of the auxiliary task of poker face generation, the proposed PF-ViT focus on the entire face, indicating PF-ViT captures and emphasizes global, emotion-relevant features.  
	   }
	\label{fig:attention}
\end{figure}

\textbf{Visualization}.  We show the attention maps of our different models based on ViT-B in Fig. \ref{fig:attention}, where SL-ViT and MAE-ViT represent the plain ViTs pre-trained using ImageNet1K data and AffectNet-270K data, respectively, which are selected from Table \ref{ViTbaseline}.
The attention maps are generated using \emph{GradCAM++} and resized to match the original input size.  
It can be concluded that the proposed PF-ViT pays more attention to the global information and captures more expressional details compared to SL-ViT and MAE-ViT, allowing for a comprehensive understanding of different facial expressions. Moreover, this characteristic helps alleviate the issue of overfitting specific facial regions cased by the dataset bias, enhancing the model's generalization capability.

\begin{figure}[htp]
	\centering
	\includegraphics[width=0.9\linewidth]{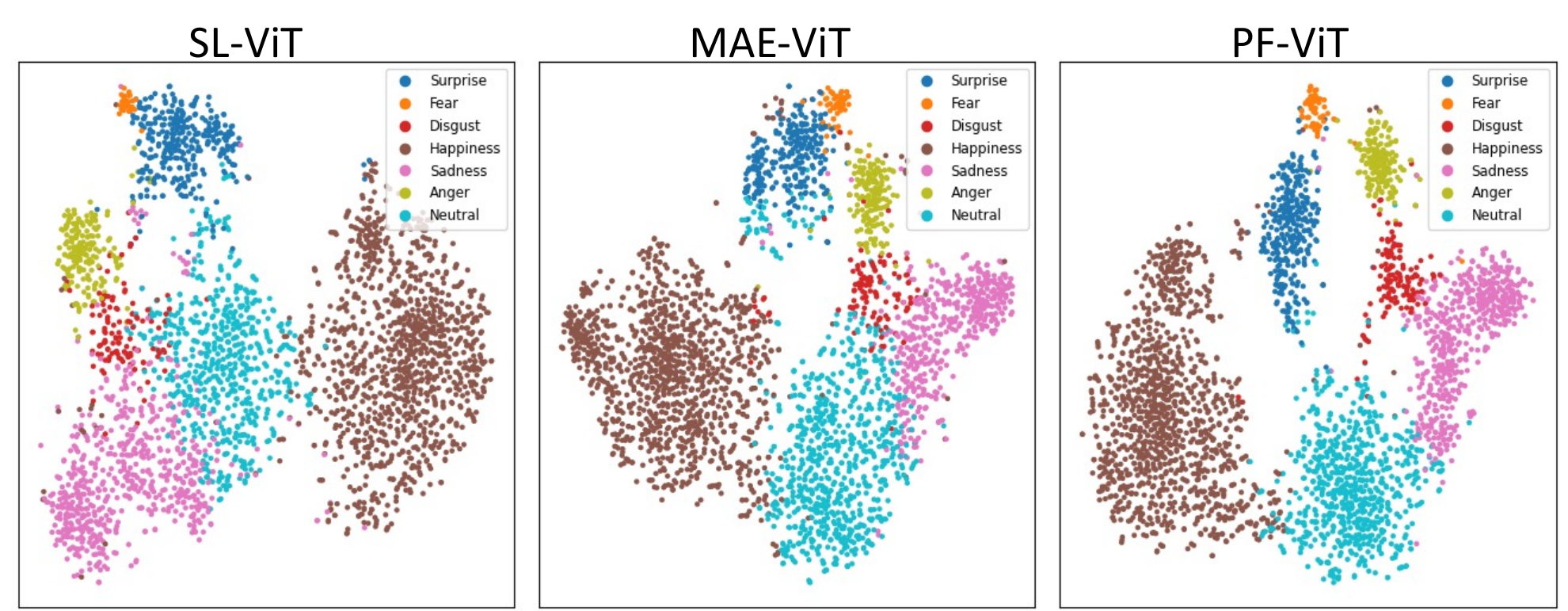}
	\caption{High-level feature visualization for SL-ViT, MAE-ViT and PF-ViT on the RAF-DB tesing set using t-SNE. The proposed PF-ViT is able to learn more discriminative expression features in the separated emotional subspace, compared to SL-ViT and MAE-ViT. The FER models above utilize ViT-B for fair.}
	\label{fig:tsne}
\end{figure}

We further visualize the high-level features of the RAF-DB dataset using t-SNE in Fig. \ref{fig:tsne}. The feature embeddings are extracted from the penultimate layers before the classification heads of SL-ViT, MAE-ViT and PF-ViT, respectively. From the visualization, it can be conclude that our PF-ViT has learned more clusterable and discriminative expression features from the disentangled emotional component $\boldsymbol{v_{e}}$, leading to more accurate FER performance.

\subsection{Comparisons with the State of the Art}
\textbf{Comprehensive comparison}. In Table \ref{Networkcompare} and Fig. \ref{fig:FLOPs}, we compare our MAE-ViT and PF-ViT, both utilizing the ViT-B backbone as the \emph{default option}, with the existing SOTA FER methods. The comparison involves metrics such as accuracy, computational complexity during testing, and model size. 
Equipped with MAE representations learn on the introduced AffectNet-270K datast, vanilla ViTs achieve competitive FER performance without the need for large-scale upstream datasets for representation learning. As shown in  Fig. \ref{fig:FLOPs}, our MAE-ViT baseline, based on ViT-T and comprising only 5.6M parameters, is even comparable to the earlier advanced FER methods. For reference, the widely used lightweight ResNet-18 backbone, which is included in Table \ref{Networkcompare}, contains approximately 12M parameters. 

During inference for the FER task only, the image generator $G$ of PF-ViT can be removed. Thus, $G$ is not  taken into account when computing the computational complexity of our PF-ViT, as shown in Fig. \ref{fig:FLOPs}.
Based on the results presented in Table \ref{Networkcompare}, our \textbf{\emph{PF-ViT has achieved a good trad-off between accuracy and computational cost, seting a new state of the art}} with the highest accuracy numbers of 92.07\%, 67.23\%, 64.10\% and 91.16\%  on the RAF-DB, AffectNet-7, AffectNet-8 and FERPlus testing tests, respectively. 
Compared to the pre-trained MAE encoder detailed in Table \ref{MAEConfig}, the additional classification head $C$ and token separator $S$ in PF-ViT only contribute a small increase in the number of model parameters  ($89.7\text{M}  - 85.05\text{M} = 4.65\text{M}$) or computational cost  ($9.0\text{G}  - 8.6\text{G} = 0.4\text{G}$ FLOPs). 
Finally, our results demonstrate the high potential of plain ViT models for the task of FER.

\begin{table}
	\caption{Comparison of our framework (PF-MAG in Fig. \ref{fig:PFViT}) with several SOTA FER methods in terms of model size (number of parameters) and computational complexity (FLOPs) with the input size of 224$\times$224 during training. ViT-B is our default image encoder.}
	\centering
	\resizebox{0.8\columnwidth}{!}{
		\smallskip\begin{tabular}{lccc}
			\hline training framework & backbone & \#params & FLOPs \\
			\hline SCN \cite{wang2020suppressing} & ResNet-18 & $11.2 \mathrm{M}$ & $1.82 \mathrm{G}$ \\
			RUL \cite{zhang2021relative} & ResNet-18 & $14.39 \mathrm{M}$ & $4.95 \mathrm{G}$ \\
			MA-Net \cite{zhao2021learning} & ResNet-18 & $63.5 \mathrm{M}$ & $3.67 \mathrm{G}$ \\
			AMP-Net \cite{liu2022adaptive} & ResNet-34 & $105.7 \mathrm{M}$ & $4.73 \mathrm{G}$ \\
			MRAN \cite{chen2023multi} & Hybrid ViT & $60.5 \mathrm{M}$ & $3.89 \mathrm{G}$ \\
			VTFF \cite{ma2021facial} & Hybrid ViT & $51.8 \mathrm{M}$ & $6.08 \mathrm{G}$ \\
			\hline \textcolor{lightgray}{MAE \cite{he2022masked} (Pre-training)} &\textcolor{lightgray}{Pure ViT} &\textcolor{lightgray}{$111.65 \mathrm{M}$} &\textcolor{lightgray}{$9.4 \mathrm{G}$} \\
			PF-MAG (Ours) & Pure ViT & $145.1 \mathrm{M}$ & $15.2 \mathrm{G}$ \\
			\hline
			\end{tabular}}
	\label{traing_cost}
\end{table}

\begin{table}
	\caption{Comparison of inference speed during testing. We run the following open-source models on either a single Nvidia 3090 GPU or a 2080Ti GPU with the input size of 224$\times$224. The speed numbers are obtained using FP32 (single-precision) inference. It is worth noting that batch size practically affects testing speed.}
	\centering
	\resizebox{1\columnwidth}{!}{
		\smallskip\begin{tabular}{c|cccc}
			\hline
			& \multicolumn{4}{c}{\textbf{FPS on a single 2080Ti / 3090 GPU}} \\
			\hline
			\diagbox{model}{batch size} & 1 & 6 & 16 & 32  \\
			\hline
			SCN \cite{wang2020suppressing}   & 29.8 / 35.6  & 127.4 / 156.4 & 292.2 / 343.5 & 395.0 / 465.2   \\
			RUL  \cite{zhang2021relative}   & 23.1 / 28.3  & 113.0 / 139.4 & 253.9 / 323.5 & 338.7 / 421.1   \\
			MA-Net \cite{zhao2021learning}   & 20.5 / 25.2  & 90.1 / 114.9  & 201.6 / 244.7 & 274.6 / 329.5   \\
			AMP-Net \cite{liu2022adaptive}   & 18.3 / 24.3  & 82.6 / 107.2  & 185.9 / 231.6 & 261.1 / 338.7   \\
			\textbf{PF-ViT, w/o ($G$, $D$)}  & 33.4 / 41.3  & 124.4 / 148.5 & 285.8 / 352.4 & 389.9 / 490.8   \\
			\hline
		\end{tabular}
	}
	\label{tesing_speed}
\end{table}

\textbf{Comparision on model parameters and inference speed}. During the training phase, the inclusion of the cross-fusion module $MCA$, generator $G$, discriminator $D$ contributes to increased training cost. We compare our entire GAN framework, i.e., PF-MAG, with other SOTA works in Table \ref{traing_cost}, focusing on model parameters and computational complexity during the training stage. For fair comparison, all trainable components of the frameworks in the table are taken into account. Considering we are the first to apply pure ViTs for FER and have achieved the best accuracy performance, our framework's parameter size (145.1M) and computational cost (15.2 GFLOPS) are still tolerable to some extent, despite this shortcoming. 
Fortunately, Our final FER model comprising only ($E$, $S$, $C$), with 89.7M parameters and 9.0G FLOPs as  illustrated in Fig. \ref{fig:FLOPs}, can obtain high testing speed on both 2080Ti and 3090 GPUs in practical settings. Table \ref{tesing_speed} presents the practical speed metrics. 
Keep in mind that FLOPs (Floating Point Operations) does not always accurately depict the actual speed of running on a GPU.

\section{Conclusion}    \label{sec:our-conclusion}
In this paper, we firstly demonstrate that vanilla ViTs with MAE pre-training are able to achieve competitive performance for the task of FER, without the need for extra labeled data.
Then, we propose PF-ViT based on vanilla Transformer backbones to simultaneously perform facial expression classification and poker face generation, although block artifacts may exist in the generated images. Our  quantitative and qualitative results suggest that the introduced auxiliary task of poker face generation guides PF-ViT to capture fine-grained facial features holistically, leading to improved FER performance. PF-ViT is trained using unpaired images in a GAN framework, and during testing, its image generator can be removed to avoid 
additional computational cost. Compared to previous methods, our method has achieved a good balance between accuracy and inference cost, setting a new state of the art on in-the-wild FER datasets.
In future work, 
it would be interesting to investigate the feasibility of transfering expressions to tackle the challenge of class imbalance in real-world FER datasets by extending our method, as the emotional component can be effectively separated and plausible emotionless faces can be synthesized.

\appendices   

\setcounter{table}{0}   
\setcounter{figure}{0}
\renewcommand{\thetable}{A\arabic{table}}
\renewcommand{\thefigure}{A\arabic{figure}}

\section{Failure examples produced by Our PF-ViT}

\begin{figure}[!h]
	\centerline{\includegraphics[width=0.95\linewidth]{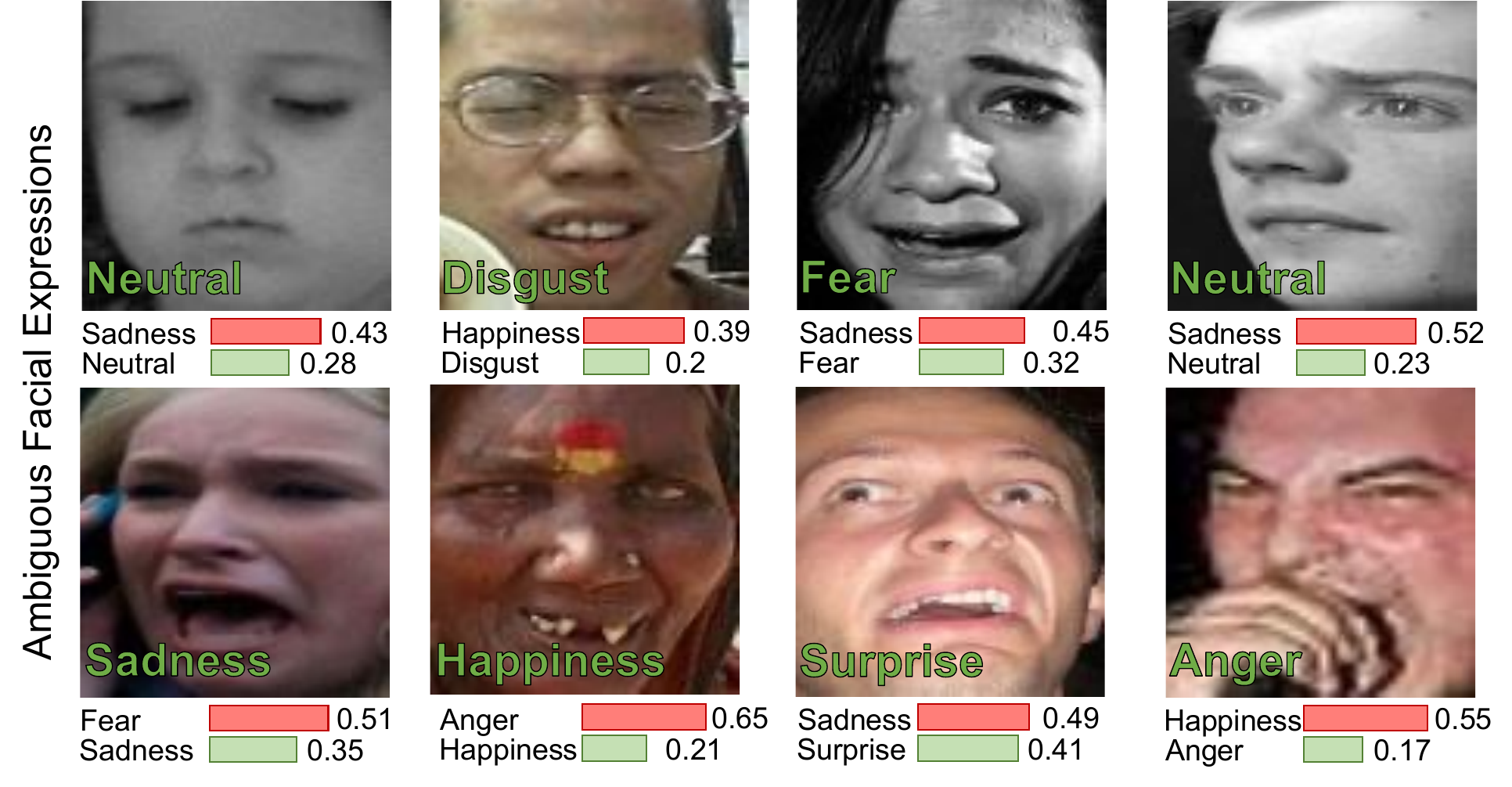}}
		\caption{Examples of incorrect predictions made by the proposed PF-ViT, which can be attributed to the inherent ambiguity of facial expressions. 
		}
	\label{fig:wrong_predictions}
\end{figure}

Some ambiguous facial expressions, which are difficult to classify, are shown in Fig. \ref{fig:wrong_predictions}. The inference scores for the top-2 predicted classes are displayed below the images and  the \textcolor{green}{ground-truth} labels are annotated within the images.

\section{Multi-Task MAE for Simultaneous FER and Poker Face Generation} \label{sec:multi_task_MAE}
\setcounter{table}{0}  
\setcounter{figure}{0}
\renewcommand{\thetable}{B\arabic{table}}
\renewcommand{\thefigure}{B\arabic{figure}}

\begin{figure}[!h]
	\centerline{\includegraphics[width=1\linewidth]{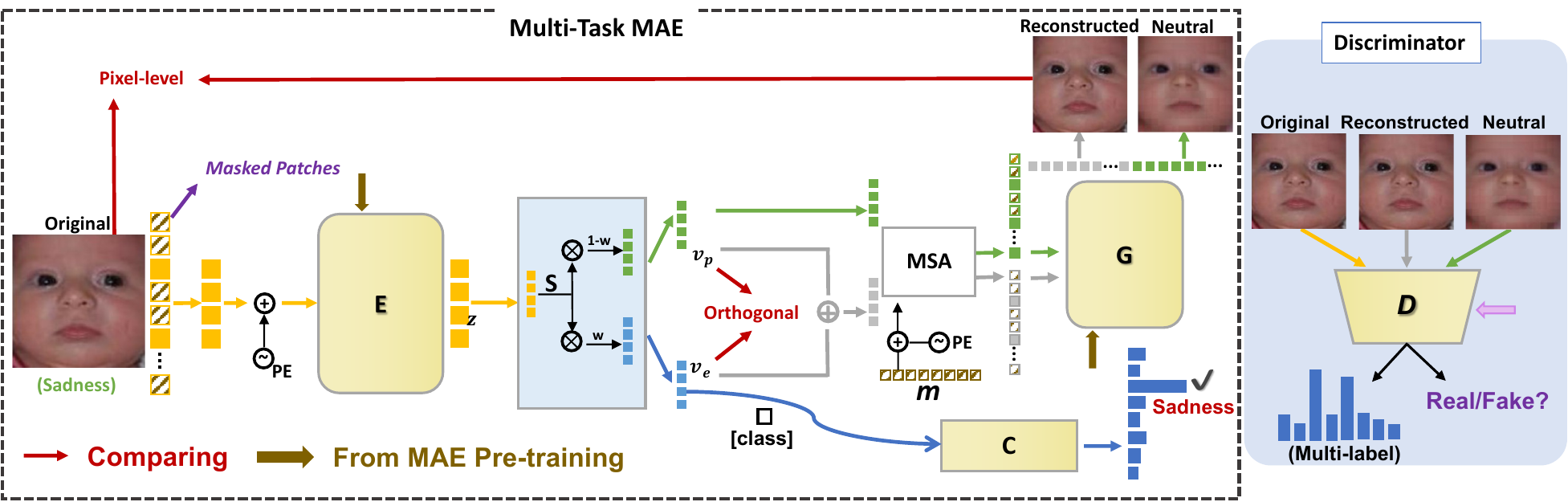}}
	   \caption{Structure of Multi-Task MAE.}
	\label{fig:Multitask-MAE}
\end{figure}

Here, we present a variation of PF-ViT called Multi-Task MAE, which attempts to follow the computation flow of Masked Autoencoders (MAEs) \cite{he2022masked}. The structure of Multi-Task MAE is illustrated in Fig. \ref{fig:Multitask-MAE}. In PF-ViT, all image tokens from the input are utilized, and a multi-head cross-attention block (denoted as ``Cross Fusion'' in Fig. \ref{fig:PFViT}) is employed to process the visual tokens and mask tokens (obtained from our earlier MAE pre-training). By contrast, Multi-Task MAE only sees the visible subset of the orgianl image patches and utilizes a multi-head self-attention block (denoted as ``MSA'' in Fig. \ref{fig:Multitask-MAE}) to process the visible tokens and mask tokens, obtaining a latent representation that is more aligned with the practice in MAEs  \cite{he2022masked}, thereby narrowing the gap between the upstream MAE pre-training and our specific application.

\begin{figure}[!h]
	\centerline{\includegraphics[width=0.75\linewidth]{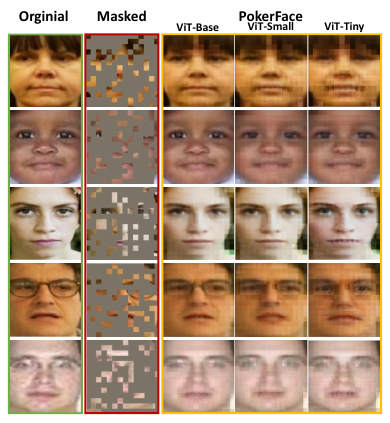}}
		\caption{Qualitative results of the auxiliary task of poker face generation, which are produced by Multi-Task MAE models using different ViT encoders, i.e., ViT-Base, ViT-Small and ViT-Tiny. Here, we set the mask ratio, $r_{mask}=0.75$ during training and $r_{mask}=0$ during testing. The poker face images generated here are not as satisfactory as those produced by PF-ViT, which utilizes all the image patches.}
	\label{fig:MAEauxiliary}
\end{figure}

We train Multi-Task MAE with random masking on the input image. To investigate the influence of different mask ratio values, we have tried various options, namely 0 (no mask tokens), 0.3, 0.5, and 0.75.  The results on the RAF-DB testing set are summarized in Table \ref{MultitaskMAE-mask-ratio}, demonstrating that a smaller mask ratio leads to higher FER accuracy during testing, while a moderate value of $r_{mask}=0.5$ performs better during training. These findings emphasize the significance of choosing an appropriate mask ratio for Multi-Task MAE. Additionally, we show some qualitative results in Fig. \ref{fig:MAEauxiliary}.

\begin{table}[!h]
	\caption{Different mask ratio configurations for Multi-Task MAE during training/testing on RAF-DB dataset. We report the performance on the testing set.
	Here, ViT-Base is used as the encoder. It is important to note that the synthesis of realistic poker faces is not possible without the presence of mask tokens in both Multi-Task MAE and our PF-ViT models.} \smallskip
	\centering
	\resizebox{1\columnwidth}{!}{
		\smallskip\begin{tabular}{c|ccccc}
			\hline \textit{train/test $r_{mask}$} & $0 / 0$ & $0.3 / 0$ & $0.3 / 0.3$ & $\mathbf{0 . 5 / 0}$ & $0.5 / 0.5$ \\
			\hline accuracy (\%) & 90.23 & 89.31 & 86.95 & $\mathbf{9 0 . 3 2}$ & 87.42 \\
			\hline \textit{train/test $r_{mask}$} & $0.75 / 0$ & $0.75 / 0.75$ & $0.75 / 0.5$ & $0.75 / 0.3$ & \\
			\hline accuracy (\%) & 89.67 & 86.03 & 86.56 & 86.21 & \\
			\hline
			\end{tabular}}
		\begin{tablenotes}
			\item $\dagger$ The \textit{mask ratio}, denoted by $r_{mask}$, is defined in Eqn. \ref{eq:r_mask}. An MAE model pre-trained with a mask ratio can be appliedto input images with a different mask ratio during testing.
		\end{tablenotes}
	\label{MultitaskMAE-mask-ratio}
	\end{table}

Based on the results shown in Table \ref{MultitaskMAE-mask-ratio} and Fig. \ref{fig:MAEauxiliary}, we can conclude that PF-ViT (with an accuracy of 92.07\%  on the RAF-DB testing set), which utilizes all the image patches to capture all the facial details, significantly outperforms Multi-Task MAE in terms of the facial expression classification accuracy and the image quality of synthesized poker faces. Multi-Task MAE with the best accuracy is even inferior to our MAE-ViT baseline, which utilizes the ViT-B backbone (91.07\%).



	

\bibliographystyle{IEEEtran}
\bibliography{IEEEabrv,mybib}

\end{document}